\documentclass[runningheads]{llncs}
\usepackage{graphicx}
\usepackage{graphicx}
\usepackage{subfigure}
\usepackage{multirow}
\usepackage{amsmath}

\begin{document}
\title{Combining Improvements for Exploiting Dependency Trees in Neural Semantic Parsing}

%
%
\author{Defeng Xie\inst{1}\orcidID{0000-0001-8036-6130} \and 
Jianmin Ji\inst{1}\orcidID{0000-0002-1515-0402} \and
Jiafei Xu\inst{1}\orcidID{0000-0002-6793-0080} \and
Ran Ji\inst{1}\orcidID{0000-0003-3735-1421}
 }
%
%
\institute{University of Science and Technology of China, Heifei, China\\
	\email{jianmin@ustc.edu.cn,\{ustcxdf,xujiafei,jiran\}@mail.ustc.edu.cn}}
\maketitle 
%
%
\begin{abstract}
The dependency tree of a natural language sentence can capture the interactions between semantics and words.
However, it is unclear whether those methods which exploit such dependency information for semantic parsing can be combined to achieve further improvement and the relationship of those methods when they combine.
In this paper, we examine three methods to incorporate such dependency information in a Transformer based semantic parser and empirically study their combinations.
We first replace standard self-attention heads in the encoder with parent-scaled self-attention (PASCAL) heads, i.e., the ones that can attend to the dependency parent of each token.
Then we concatenate syntax-aware word representations (SAWRs), i.e., the intermediate hidden representations of a neural dependency parser, with ordinary word embedding to enhance the encoder. 
Later, we insert the constituent attention (CA) module to the encoder, which adds an extra constraint to attention heads that can better capture the inherent dependency structure of input sentences.
Transductive ensemble learning (TEL) is used for model aggregation, and an ablation study is conducted to show the contribution of each method.
Our experiments show that CA is complementary to PASCAL or SAWRs, and PASCAL + CA provides state-of-the-art performance among neural approaches on ATIS, GEO, and JOBS.

\end{abstract}
\section{Introduction}

Semantic parsing is the task of mapping natural language sentences into target formal representations, which is crucial for many natural language processing (NLP) applications. 
With the rapid development of deep learning, various neural semantic parsers~\cite{dong2016language,dong2018coarse,jia-liang-2016-data,shaw2019generating,sun2020treegen} have been implemented based on sophisticated sequence-to-sequence (seq2seq) models~\cite{cho2014learning} with Transformer~\cite{vaswani2017attention}.
Note that syntax information of natural language sentences can be used as clues and restrictions for semantic parsing. 
In this paper, we focus on incorporating syntax information from dependency trees of natural language sentences in neural semantic parsers.

The dependency tree~\cite{dependencytree1,dependencytree2} of a natural language sentence shows which words depend on which other words in a tree structure that can capture the interactions between the semantics and natural language words.
In specific, the dependency tree can be considered as the explicit structure prior to predicting corresponding semantic structures in semantic parsing. 
Then the encoder of a seq2seq based semantic parser can be enhanced by information from dependency trees.
On the other hand, dependency trees can be efficiently generated by existing parsers, like Stanford Parser~\cite{stanfordparser}, with promising results.
\cite{xu2018exploiting} encodes such dependency trees in a graph-based neural network for semantic parsing and achieves a great improvement in performance, which indicate potential advantages of incorporating dependency trees in semantic parsers.

it is unclear whether those methods which exploit such dependency information for semantic parsing can be combined for further improvement.
In this paper, we examine three such methods for a Transformer encoder of a seq2seq based semantic parser and empirically study their combinations. 
In specific, we first follow the idea of parent-scaled self-attention (PASCAL)~\cite{enhancingMTwithDASA}, which replaces standard self-attention heads in the encoder with ones that can attend to the dependency parent of each token. 
We also concatenate syntax-aware word representations (SAWRs)~\cite{syntaxenhancedNMT}, i.e., the intermediate hidden representations of a neural dependency parser, with ordinary word embedding to enhance the Transformer encoder. 
At last, we insert constituent attention (CA) module~\cite{Wang2019TreeTI} to the Transformer encoder, which adds an extra constraint to attention heads to follow tree structures that can better capture the inherent dependency structure of input sentences. 
We also aggregate multiple models of these methods for inference following transductive ensemble learning (TEL)~\cite{transductiveEL}.

We first implement a baseline semantic parser that is based on a simple seq2seq model consisting of a 2-layer Transformer encoder and a 3-layer Transformer decoder.
Then we evaluate the performance of the above three methods and their combinations on ATIS, GEO, and JOBS datasets. We also evaluate aggregated versions of these methods by TEL.
The experimental results show that the combination of PASCAL and CA provides state-of-the-art performance among neural approaches, which can also be easily implemented. 

The main contributions of this paper are: 
\begin{itemize}
	\item We introduce three methods by applying PASCAL, SAWRs, or CA to incorporate dependency trees in the Transformer encoder of a seq2seq semantic parser. We show that all three methods can improve the performance.
	\item We evaluate the combinations of the three methods and show that they can be fruitfully combined with better performance. The result show that CA is complementary to PASCAL or SAWRs.
	\item We implement TEL for our models. We show that TEL is effective for these improvements on semantic parsing.
	\item We implement the combination of PASCAL and CA based on a simple seq2seq semantic parser. We show that this parser can be implemented easily and achieves state-of-the-art performance among neural approaches on ATIS, GEO, and JOBS datasets.  
\end{itemize}

\section{Related Work}

Neural semantic parsing has achieved promising results in recent years, where various sophisticated seq2seq models have been applied. 
Many works focus on integrating the syntax formalism of target representation into the decoder of the seq2seq model. 
For instance, hierarchical tree decoders are applied in~\cite{dong2016language,sun2019grammar} to take into account the tree structure of the logical expression. Sequence-to-tree (seq2tree) model~\cite{dong2016language} updates the decoder by hierarchical tree-long short-term memory (Tree-LSTM), which helps the model to utilize the hierarchical structure of logical forms. 
\cite{Yin2018TRANXAT,sun2019grammar,sun2020treegen} first map a natural language sentence into an abstract syntax tree (AST), then serve it as an intermediate meaning representation and incorporate it with grammar rules, finally parse the AST to the corresponding target logic form.

On the other hand, there are few works on incorporating syntax information of input natural language sentences to the encoder. 
Graph-to-sequence (graph2seq) model~\cite{xu2018exploiting} constructs a graph encoder to exploiting rich syntactic information for semantic parsing.

It has shown that syntax information of input natural language sentences can be helpful for the encoder in neural machine translation (NMT) tasks~\cite{bahdanau2015neural}. 
In specific, \cite{enhancingMTwithDASA} places parent-scaled self-attention (PASCAL) heads, which can attend to the dependency parent of each token, in the Transformer encoder to improve the accuracy of machine translation.
\cite{syntaxenhancedNMT} concatenates syntax-aware word representations (SAWRs), i.e., the intermediate hidden representations of a neural dependency parser, with ordinary word embedding to enhance the Transformer encoder.
\cite{Wang2019TreeTI} introduces constituent attention (CA) module, which adds an extra constraint to attention heads to follow tree structures that can better capture the inherent dependency structure of input sentences. 
In this paper, we examine these ideas in semantic parsing and empirically study their combinations.

\section{Three Improvements}

In this section, we specify three improvements to incorporate dependency trees in the Transformer encoder of a seq2seq semantic parser.

\begin{figure}[htp]
	\centering
	\vspace{-0.35cm}
	\subfigtopskip=2pt
	\subfigbottomskip=0pt
	\subfigcapskip=-30pt
	\includegraphics[width=0.6\columnwidth]{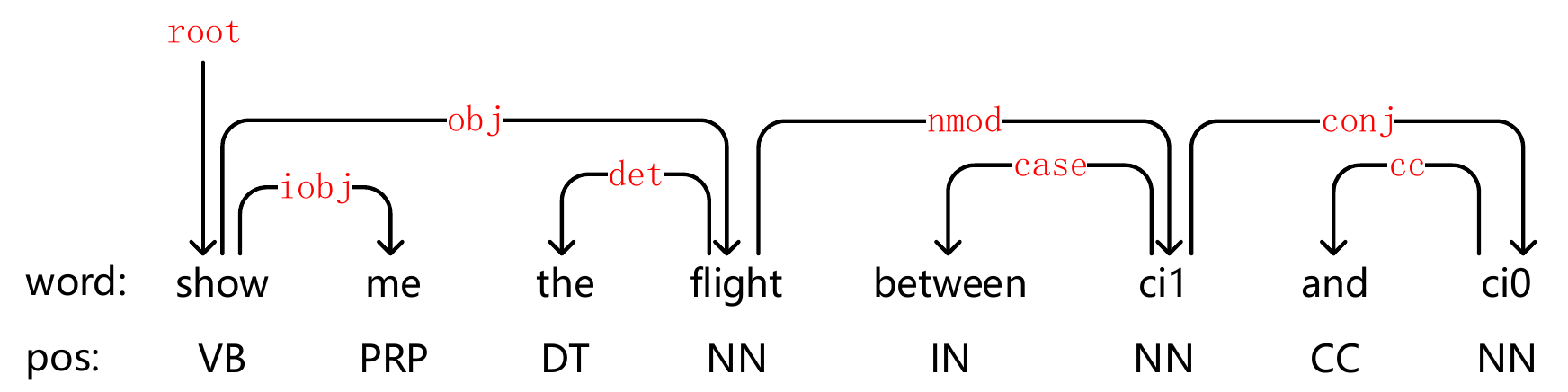}
	\caption{A dependency tree}
	\label{dependencytree}
\end{figure}

As illustrated in Figure~\ref{dependencytree}, a dependency tree describes the structure of the sentence by relating words in binary relations, which can be efficiently generated by corresponding parsers, like Stanford Parser\footnote{\url{https://nlp.stanford.edu/software/stanford-dependencies.shtml}.}.
\cite{dependencytree1,dependencytree2} have shown that dependency trees can be used to construct target logical forms for semantic parsing. 
In this paper, we focus on exploiting information from structures of these dependency trees to enhance the encoder of a neural semantic parser. Note that, we ignore the labels of corresponding dependency relations here, like `obj', `case', and `conj' in the example.

\begin{figure*}[t]
	\centering
	\vspace{-0.35cm}
	\subfigtopskip=2pt
	\subfigbottomskip=0pt
	\subfigcapskip=-10pt
	\subfigure[PASCAL]
	{
		\begin{minipage}[b]{.30\linewidth}
			\centering
			\includegraphics[width=\linewidth]{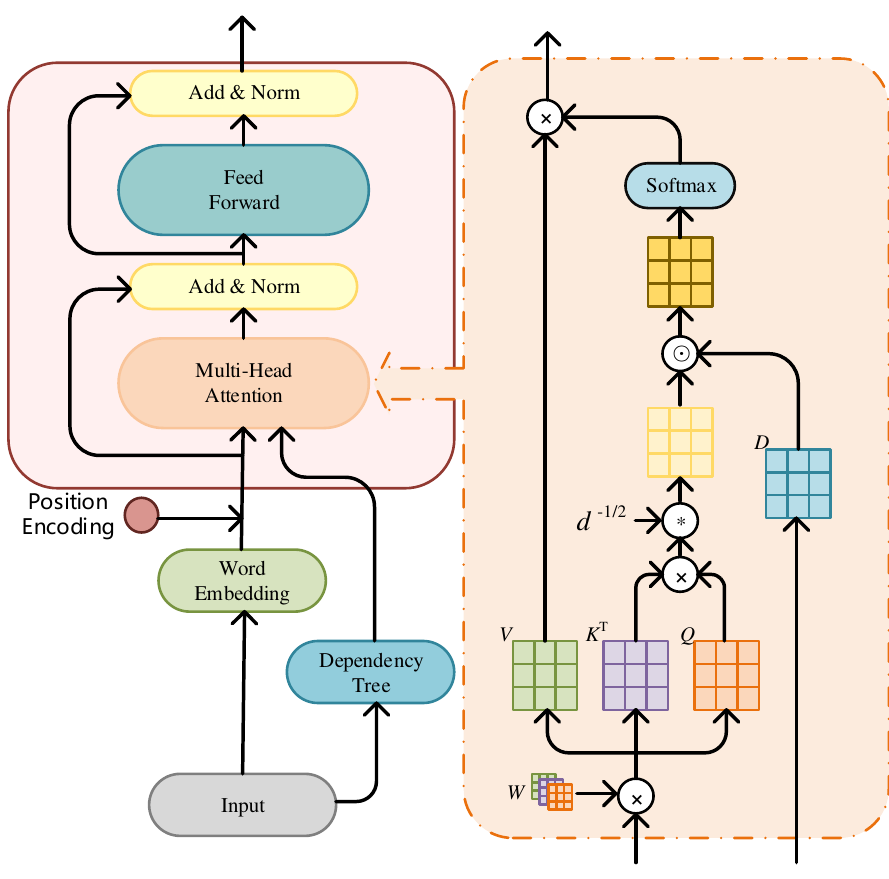}
			\label{22}
		\end{minipage}
	}\hspace{-1mm}
	\subfigure[SAWRs]
	{
		\begin{minipage}[b]{.30\linewidth}
			\centering
			\includegraphics[width=\linewidth]{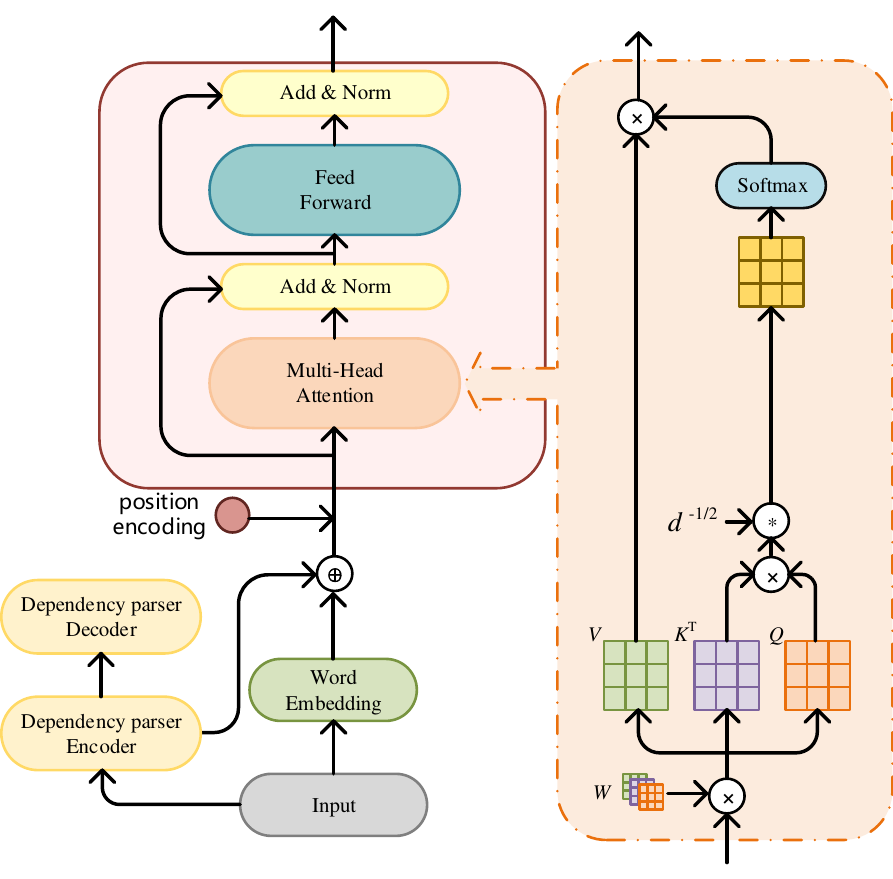}
			\label{11}
		\end{minipage}
	}\hspace{-1mm}	
	\subfigure[CA]
	{
		\begin{minipage}[b]{.30\linewidth}
			\centering
			\includegraphics[width=\linewidth]{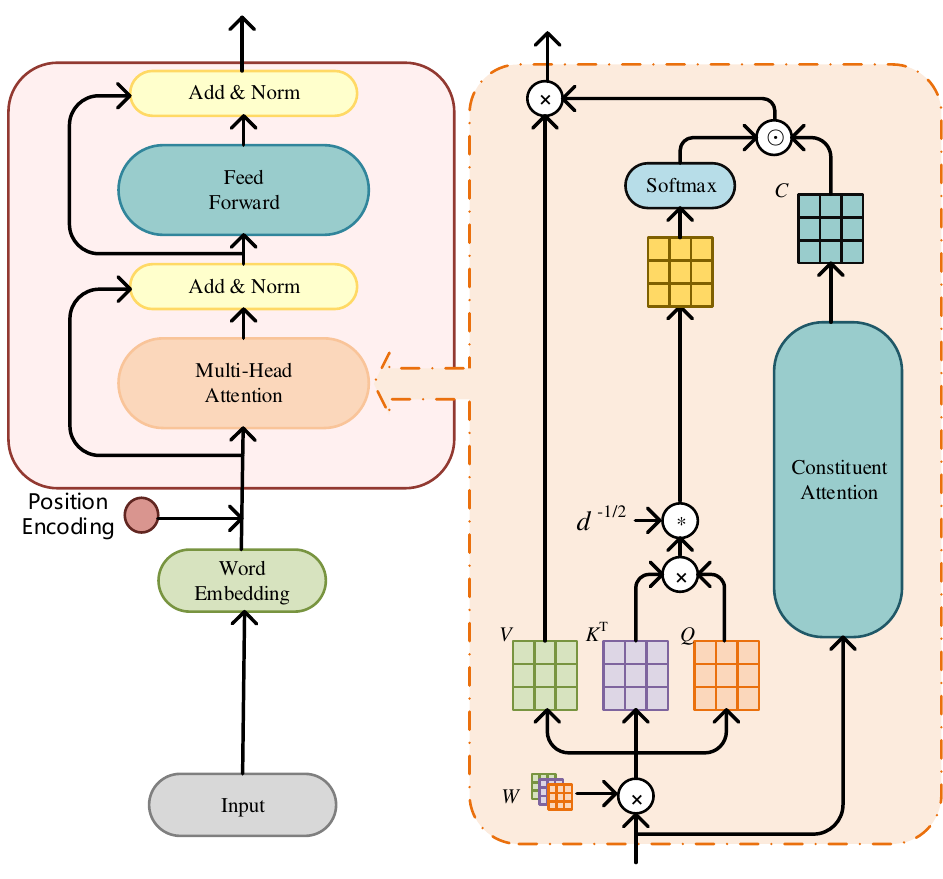}
			\label{33}
		\end{minipage}
	}
	\caption{Overviews of model structures for three improvements.}
	\label{112233}
\end{figure*}

\subsection{Parent-scaled Self-attention (PASCAL)}
\label{sec:PASCAL}

Parent-scaled self-attention (PASCAL) is first introduced in~\cite{enhancingMTwithDASA} for NMT tasks. The main idea of PASCAl is to replace standard self-attention heads in the encoder with ones that can attend to the dependency parent of each token. Here we apply the idea in semantic parsing and evaluate its effectiveness.

In specific, the standard scaled dot-product attention mechanism in Transformer is defined as follows,
\begin{equation}
\textit{Attention}(Q,K,V) = \textit{softmax}(\frac{QK^{\top}}{\sqrt{d}})V,
\end{equation}
where $Q$, $K$, $V$, $d$ denote the query matrix, the key matrix, the value matrix, and the dimension of $K$ respectively, as in~\cite{vaswani2017attention}.
We also denote $\frac{QK^{\top}}{\sqrt{d}}$ as $\textit{HeadScore}$. 

In PASCAL, $\textit{HeadScore}$ is replaced by its element-wise product with the distance matrix~$D$, which is generated from each token's dependency parent in the dependency tree by utilizing a Gaussian distribution\footnote{The detailed procedure for computing $D$ is specified in~\cite{enhancingMTwithDASA}. We omit the procedure due to the space limitation.}.
In particular, the attention mechanism used in PASCAL is defined as,
\begin{equation}\label{eq:pascal}
	\textit{Attention}(Q,K,V) = \textit{softmax}(\frac{QK^{\top}}{\sqrt{d}} \odot D)V,
\end{equation}
where $\odot$ denotes the element-wise product operation.

We implement the improvement by applying PASCAL in the Transformer encoder of a baseline seq2seq semantic parser, which consists of a 2-layer Transformer encoder and a 3-layer Transformer decoder. 
The model structure of the improved encoder is illustrated in Figure~\ref{22}, where the blue box on the right denotes the distance matrix $D$.

In this improvement, we first generate the dependency tree of an input sentence by a dependency parser. 
Then we capture the information of the dependency tree by its distance matrix~$D$.
We incorporate such dependency information in the Transformer encoder using the element-wise product of $D$ and $\textit{HeadScore}$.

\subsection{Syntax-Aware Word Representations (SAWRs)}

Given a well-trained neural dependency parser, which parses an input sentence to a dependency tree, we can obtain dependency information from its intermediate hidden representations, i.e., syntax-aware word representations (SAWRs).  
In~\cite{syntaxenhancedNMT}, such intermediate hidden representations are specified as the outputs of the BiLSTM layer in the BiAffine dependency parser~\cite{deepbiaffparser}. 

In this paper, we first train a neural dependency parser based on the model proposed in~\cite{simperparser}, which is simpler and performs better. 
We also specify SAWRs as the outputs of the BiLSTM layer in this dependency parser. 
Then we concatenate such SAWRs with ordinary word embedding to enhance the Transformer encoder for the semantic parser. 
In particular, the input of the Transformer encoder is improved from $\textrm{WE} + \textrm{PE}$ to 
$
(\textrm{SAWRs} \oplus \textrm{WE}) + \textrm{PE}
$,
where $\oplus$ denotes the concatenate operation, $\textrm{WE}$ and $\textrm{PE}$ denote the word embedding and the position encoding, respectively.

The model structure of the improved encoder is illustrated in Figure~\ref{11}, where two yellow boxes on the left corner denote the pre-trained neural dependency parser.
We will specify the training process for the dependency parser in Section~\ref{sec:impledetail}. 

In this improvement, we first train a neural dependency parser.
Then we capture the information of the dependency tree by its intermediate hidden representations, i.e., SAWRs.
We incorporate such dependency information in the Transformer encoder by concatenating SAWRs with $\textrm{WE}$.

\subsection{Constituent Attention (CA)}
\label{sec:ca}

\cite{Wang2019TreeTI} introduces Constituent Attention (CA) module, which adds an extra constraint to attention heads to follow tree structures, that can better capture the inherent dependency structure of input sentences.
Here we apply the idea in semantic parsing and evaluate its effectiveness.

In specific, the attention mechanism in Transformer is improved to 
\begin{equation}
\textit{Attention}(Q,K,V) = \left( C \odot \textit{softmax}(\frac{QK^{\top}}{\sqrt{d}})\right) V, 
\end{equation}
where $\odot$ denotes the element-wise product operation, $Q$, $K$, $V$, $d$ denote the same as above, and $C$ denotes the constituent prior generated from CA module.
In particular, $C$ is a symmetric matrix that describes the probabilities of whether two words belonging to the same constituent\footnote{The detailed procedures for constructing CA module and computing $C$ are specified in~\cite{Wang2019TreeTI}. We omit the procedures due to the space limitation.}.

The model structure of the improved encoder is illustrated in Figure~\ref{33}, where the blue round frame and the blue box on the right denote CA module. We will specify the training process for the improved model in Section~\ref{experimentprocess}.

In this improvement, we add CA module to the Transformer encoder, which introduces the constituent prior $C$ to attention heads.
Such constraint encourages the attention heads to follow tree structures, which helps the encoder to capture the inherent dependency information of input sentences.

\section{Combining Improvements}
\label{ourmethods}

In this section, we consider all possible combinations of the three improvements and integrate the combinations into a single model. We also try to further improve the performance by using the ensemble learning method, i.e., TEL.

A combination $A$ + $B$ denotes the seq2seq model that applies both improvements $A$ and $B$ in its encoder. 
In specific, both combinations PASCAL + SAWRs and SAWRs + CA can be directly implemented, as SAWRs does not affect the implementation of either PASCAL or CA.
For combinations PASCAL + CA and PASCAL + SAWRs + CA, we need to improve the attention mechanism in Transformer to
\begin{equation}
	\textit{Attention}(Q,K,V) = \left( C \odot \textit{softmax}(\frac{QK^{\top}}{\sqrt{d}}\odot D)\right) V.
\end{equation}

\begin{figure*}[t]
	\centering
	\vspace{-0.35cm}
	\subfigtopskip=2pt
	\subfigbottomskip=0pt
	\subfigcapskip=-10pt
	\subfigure[PASCAL + CA]
	{
		\begin{minipage}[b]{.40\linewidth}
			\centering
			\includegraphics[width=\linewidth]{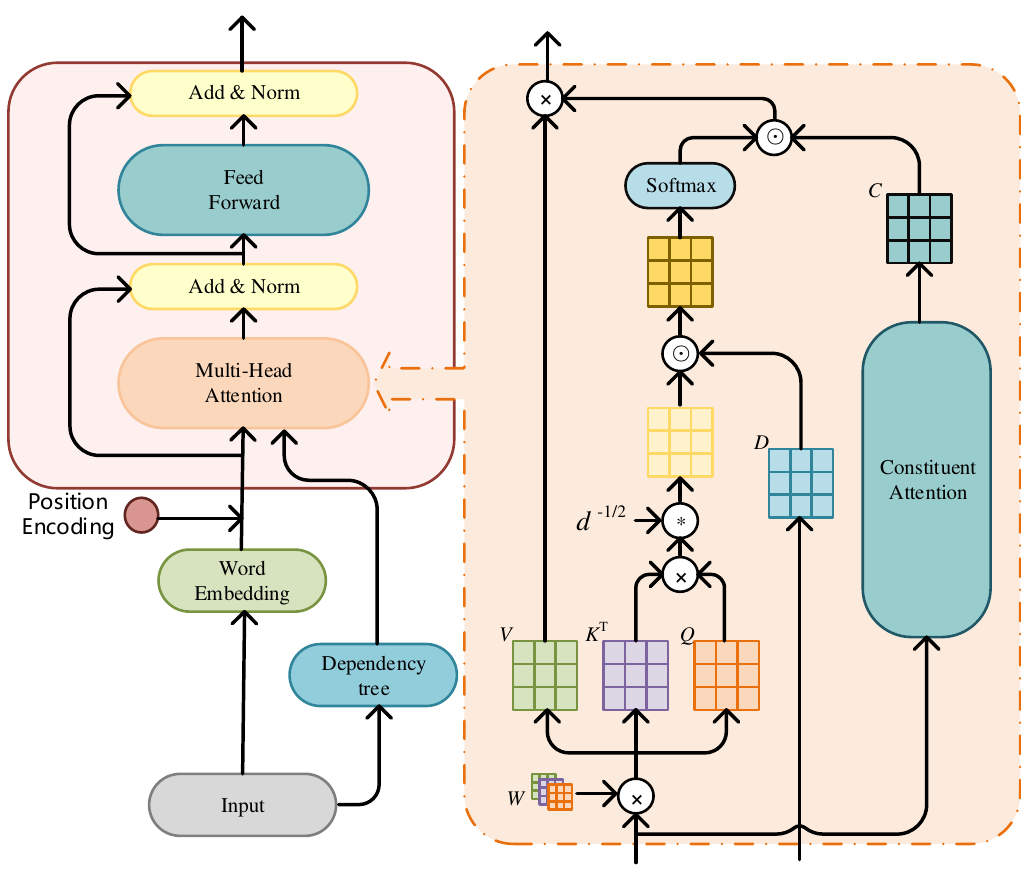}
			\label{s11}
		\end{minipage}
	}\hspace{-2mm}
	\subfigure[PASCAL+SAWRs]
	{
		\begin{minipage}[b]{.40\linewidth}
			\centering
			\includegraphics[width=\linewidth]{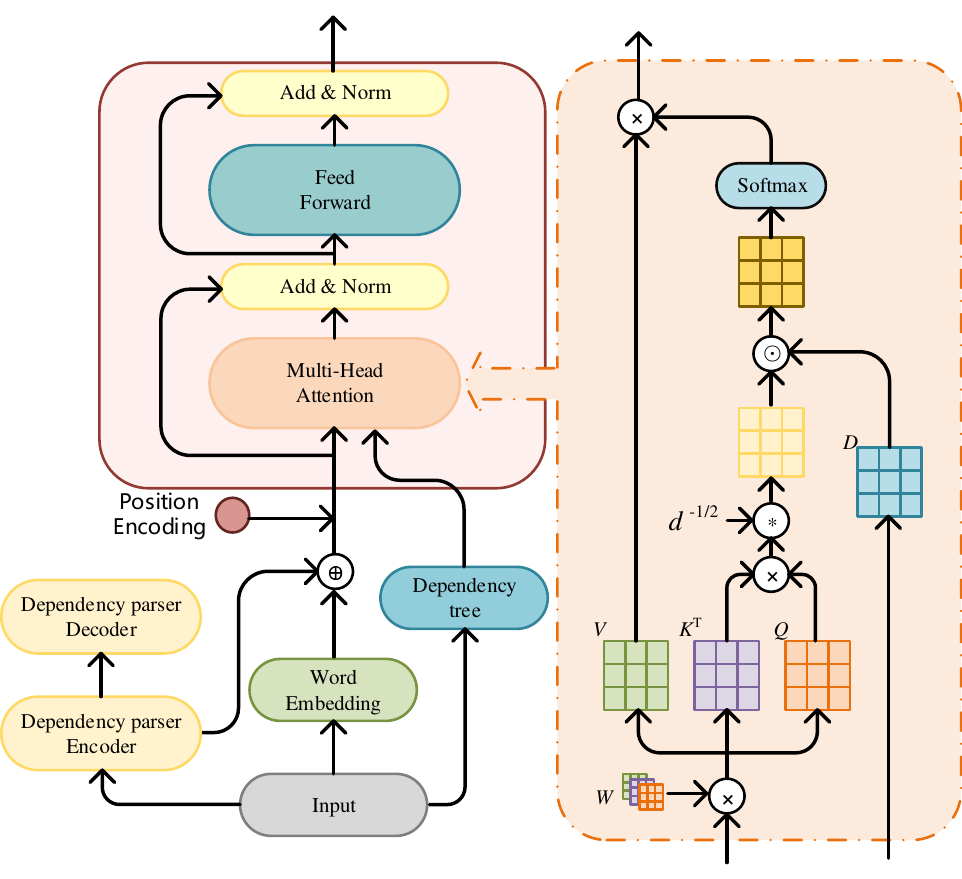}
			\label{s22}
		\end{minipage}
	}\hspace{-2mm}
	\subfigure[SAWRs+CA]
	{
		\begin{minipage}[b]{.40\linewidth}
			\centering
			\includegraphics[width=\linewidth]{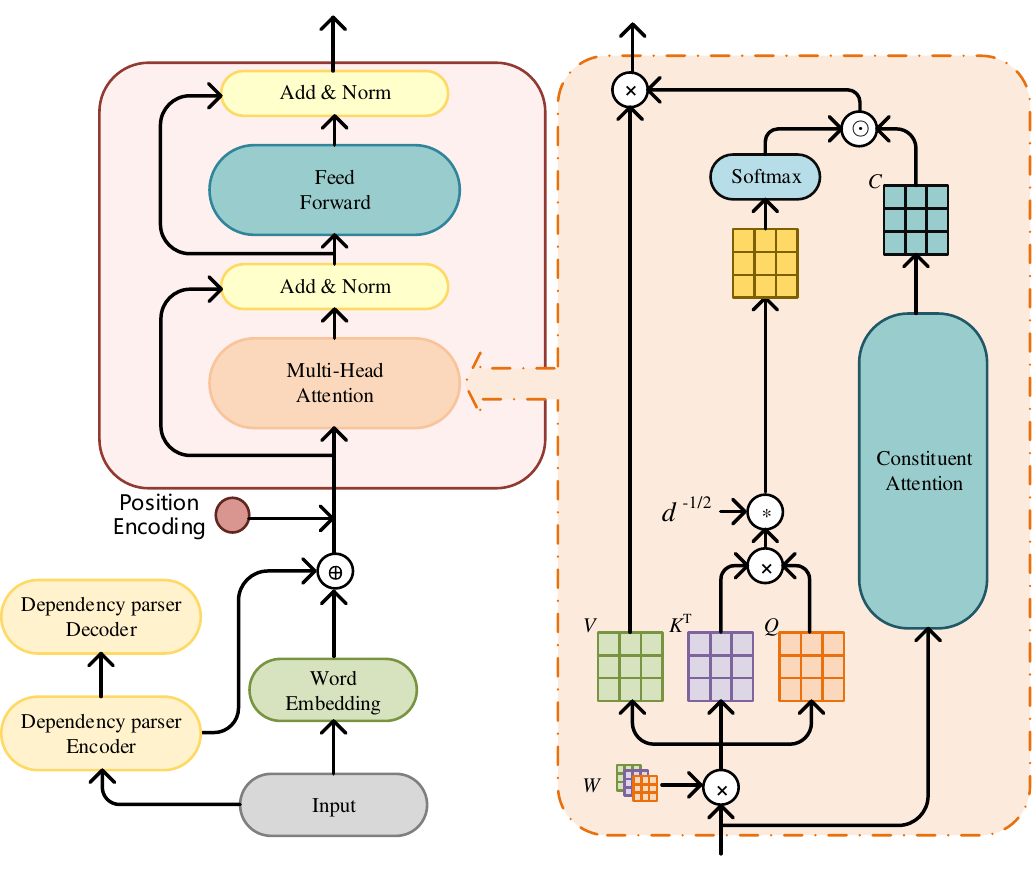}
			\label{s33}
		\end{minipage}
	}\hspace{-2mm}
	\subfigure[PASCAL+SAWRs+CA]
	{
		\begin{minipage}[b]{.40\linewidth}
			\centering
			\includegraphics[width=\linewidth]{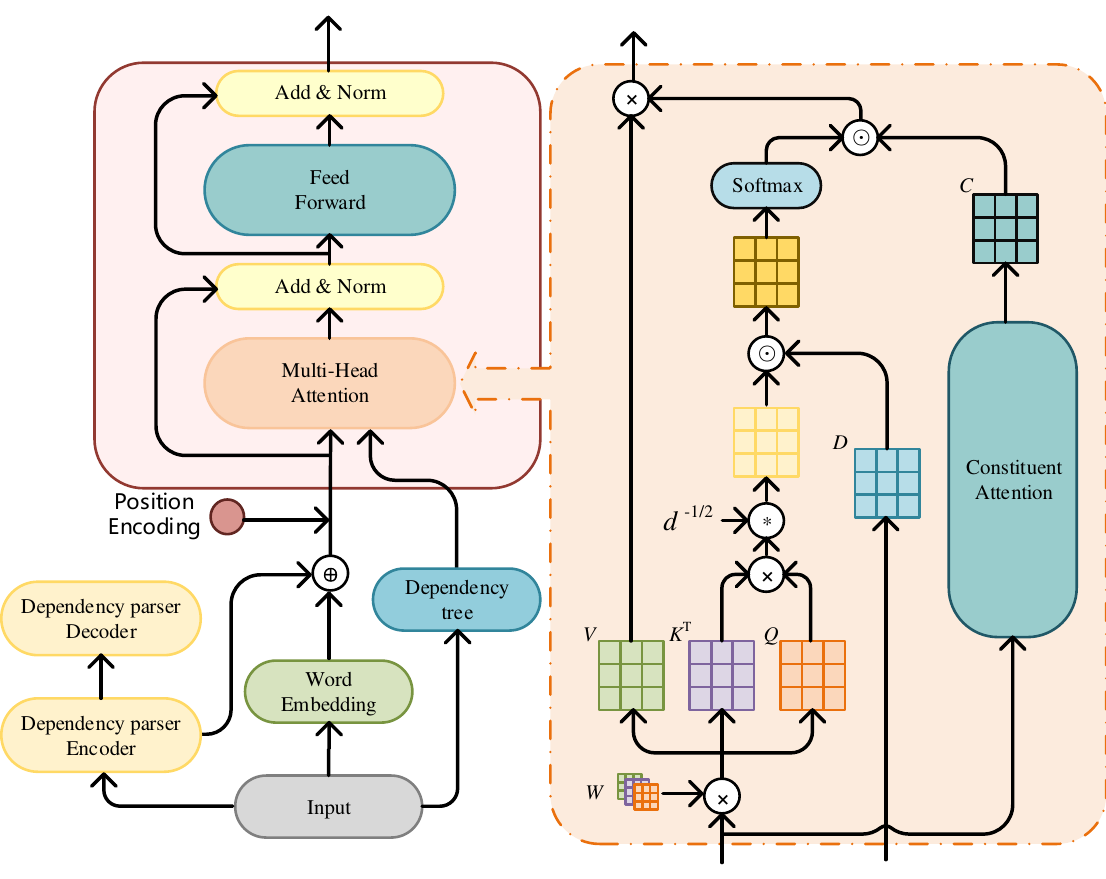}
			\label{s44}
		\end{minipage}
	}
	\caption{Overviews of model structures for combinations.}
	\label{s112233}
\end{figure*}

Model structures of the improved encoders for all combinations are illustrated in Figure~\ref{s112233}.

In this paper, we use Transductive Ensemble Learning (TEL)~\cite{transductiveEL} to aggregate multiple individual models for better performance.
Note that, TEL is applied under the transductive setting, i.e., the model can observe the input sentences in the test set.
TEL is first introduced for NMT tasks. Here we implement the idea in semantic parsing and evaluate its effectiveness.

In specific, following TEL, we first use all individual models to predict the input sentences from the validation and test sets, and construct a synthetic corpus by using these predicted results as corresponding labels.
Then we select the model with the best performance on the validation set\footnote{The datasets of GEO and JOBS are small and do not contain validation sets. Then parameters and the selected model are cross-validated on their training sets.} and fine-tune this model on the generated synthetic corpus. 
At last, we use the fine-tuned model in the inference phase.
Notice that, TEL is efficient and easy to be implemented, as only one model is selected for inference.

In the following, we use $A$ + TEL to denote the model that applies TEL for ensemble learning based on $A$.

\section{Experiments}
\label{experimentprocess}

\subsection{Datasets}

We evaluate the three improvements, their combinations, and TEL on three famous datasets for semantic parsing, i.e., ATIS, a set of 5,410 queries to a flight booking system, GEO, a set of 880 queries to a database of U.S. geography, and JOBS, a set of 640 queries to a database of job listings. 

We follow the standard train-dev-test split of these datasets and use the preprocessed version as specified in~\cite{dong2016language}. 
We also adopt Stanford CoreNLP package~\cite{stanfordparser} to do the tokenization.
Then the target formal representations of these datasets are all $\lambda$-calculus expressions here.

	

\subsection{Evaluation Metrics}

We use Exact Match~\cite{shaw2019generating} and Tree Exact Match (Tree Match)~\cite{sun2020treegen} to evaluate the performance of different models.
In particular, Exact Match computes the percentage of sentences whose predicted results are exactly the same as their labeled target logic forms, i.e., $\lambda$-calculus expressions.
However, in some cases, the order of formulas can be equivalently changed in $\lambda$-calculus expressions. For instance, the order of two formulas in conjunction can be equivalently reversed. 
Then Tree Match is introduced to avoid these spurious errors by considering the tree structures of resulting logic forms. Note that, there is little previous work using Tree Match for JOBS. Then we only use Exact Match for JOBS.

\subsection{Implementation Details}
\label{sec:impledetail}

The baseline model for the semantic parser considered here is a seq2seq model consisting of a 2-layer Transformer encoder and a 3-layer Transformer decoder.
We trained this model with the hyperparameters listed in Table~\ref{tab:hparameter}, whose parameters were chosen based on the performance of the model on the validation set for ATIS and cross-validated on the training sets for GEO and JOBS.

\begin{table}[htp]
	\centering
	\caption{Hyperparameters for the baseline model.}
	\label{tab:hparameter}
	\begin{tabular}{ll}
		\hline
		Hyperparameter & Value\\ 
		\hline
		word embedding dimension & 512 \\
		position encoding dimension & 512 \\
		Transformer head number & 8 \\
		Transformer attention dimension & 512 \\
		Transformer feed forward dimension & 2048 \\
		Transformer activation & ReLU \\
		dropout rate & 0.1 \\
		batch size & 16 \\
		learning rate & 1e-4 \\
		\hline
	\end{tabular}
	
\end{table}

\begin{table*}[htp]
	\caption{Experimental results.}
	\label{tab:result}
	\begin{center}
		\begin{tabular}{lccccc}
			\hline	
			Model &    \multicolumn {2}{c} {ATIS}& \multicolumn {2}{c} {GEO} &
			\multicolumn {1}{c} {JOBS} \\
			\hline
			\multirow{2}{*}{Evaluation Metric}  & Exact & Tree  & Exact & Tree & Exact \\
			& Match & Match & Match & Match & Match \\
			\hline
			\multicolumn {6}{c} {Pre-neural methods}\\
			\hline
			ZC07~\cite{zettlemoyer2007online} & 84.6 & - & 86.1 & - & 79.3 \\
			FUBL~\cite{kwiatkowski2011lexical} & 82.8 & - & 88.6 & - & - \\
			DCS~\cite{liang39learning} & - & - & 87.9 & - & \textbf{90.7} \\
			KCAZ13~\cite{kwiatkowski-etal-2013-scaling} & - & - & 89.0 & - & - \\
			WKZ14~\cite{wang2014morpho} & \textbf{91.3} & - & \textbf{90.4} & - & - \\
			TISP~\cite{zhao2015type} & 84.2 & - & 88.9 & - & 85.0 \\
			\hline
			\multicolumn {6}{c} {Neural methods}\\
			\hline
			Seq2Seq~\cite{dong2016language} & - & 84.2 & - & 84.6 & 87.1 \\
			Seq2Tree~\cite{dong2016language} & - & 84.6 & - & 87.1 & 90.0 \\
			JL16~\cite{jia-liang-2016-data} & 83.3 & - & 89.3$^*$ & - & - \\
			TranX~\cite{Yin2018TRANXAT} & - & 86.2 & - & 88.2 & - \\
			Coarse2fine~\cite{dong2018coarse} & - & 87.7 & - & 88.2 & - \\
			Seq2Act~\cite{chen2018sequence} & 85.5 & - & 88.2$^*$ & - & - \\
			Graph2Seq~\cite{xu2018exploiting} & 85.5 & - & 88.9 & - & \textbf{91.2} \\
			AdaNSP~\cite{zhang2019adansp} & - & 88.6 & - & 88.9 & - \\
			GNN~\cite{shaw2019generating} & \textbf{87.1} & - & \textbf{89.3} & - & - \\
			TreeGen~\cite{sun2020treegen} & - & \textbf{89.1} & - & \textbf{89.6} & - \\
			\hline
			\multicolumn {6}{c} {Our methods without TEL}\\
			\hline
			Baseline & 85.0 & 86.2 & 83.2 & 87.5 & 87.9 \\
			PASCAL & 87.5 & 88.6 & 85.0 & 88.2 & 90.7 \\
			SAWRs & 86.6 & 87.7 & 85.0 & 87.9 & 90.7 \\
			CA & 87.1 & 88.6 & \textbf{85.4} & \textbf{88.9} & 91.4 \\
			PASCAL + SAWRs & 86.8 & 87.5 & 84.3 & 88.2 & 89.3 \\
			PASCAL + CA & \textbf{88.4} & 89.1 & \textbf{85.4} & \textbf{88.9} & \textbf{92.1} \\
			SAWRs + CA & 88.0 & \textbf{89.5} & 84.3 & 87.9 & 90.7 \\
			PASCAL + SAWRs + CA & 87.7 & 89.3 & 84.3 & 87.9 & 91.4 \\
			\hline
			\multicolumn {6}{c} {Our methods with TEL}\\
			\hline
			Baseline + TEL & 87.3 & 88.4 & 84.6 & 88.2 & 88.6 \\
			PASCAL + TEL & 88.6 & 89.5 & 86.8 & 90.4 & 92.1 \\
			SAWRs + TEL & 88.4 & 89.1 & 86.8 & \textbf{90.7} & 92.1 \\
			CA + TEL & 89.1 & 90.0 & \textbf{87.1} & 90.4 & \textbf{92.9} \\
			PASCAL + SAWRs + TEL & 87.5 & 88.4 & 85.7 & 89.3 & 91.4 \\
			PASCAL + CA + TEL & \textbf{89.2} & \textbf{90.2} & \textbf{87.1} & 90.4 & \textbf{92.9} \\
			SAWRs + CA + TEL & 89.1 & \textbf{90.2} & 85.7 & 88.9 & \textbf{92.9} \\
			PASCAL + SAWRs + CA + TEL & 89.1 & 90.0 & 85.7 & 89.6 & 92.1 \\
			\hline
			\footnotesize{$^*$ Denotation Match~\cite{jia-liang-2016-data} is used.}
		\end{tabular}
	\end{center}
	
\end{table*}

\begin{figure}
	\centering
	\vspace{-0.35cm}
	\subfigtopskip=2pt
	\subfigbottomskip=0pt
	\subfigcapskip=-10pt
	\subfigure[{\scriptsize Exact Match on ATIS}]
	{
		\begin{minipage}{.18\linewidth}
			\centering
			\includegraphics[width=\linewidth]{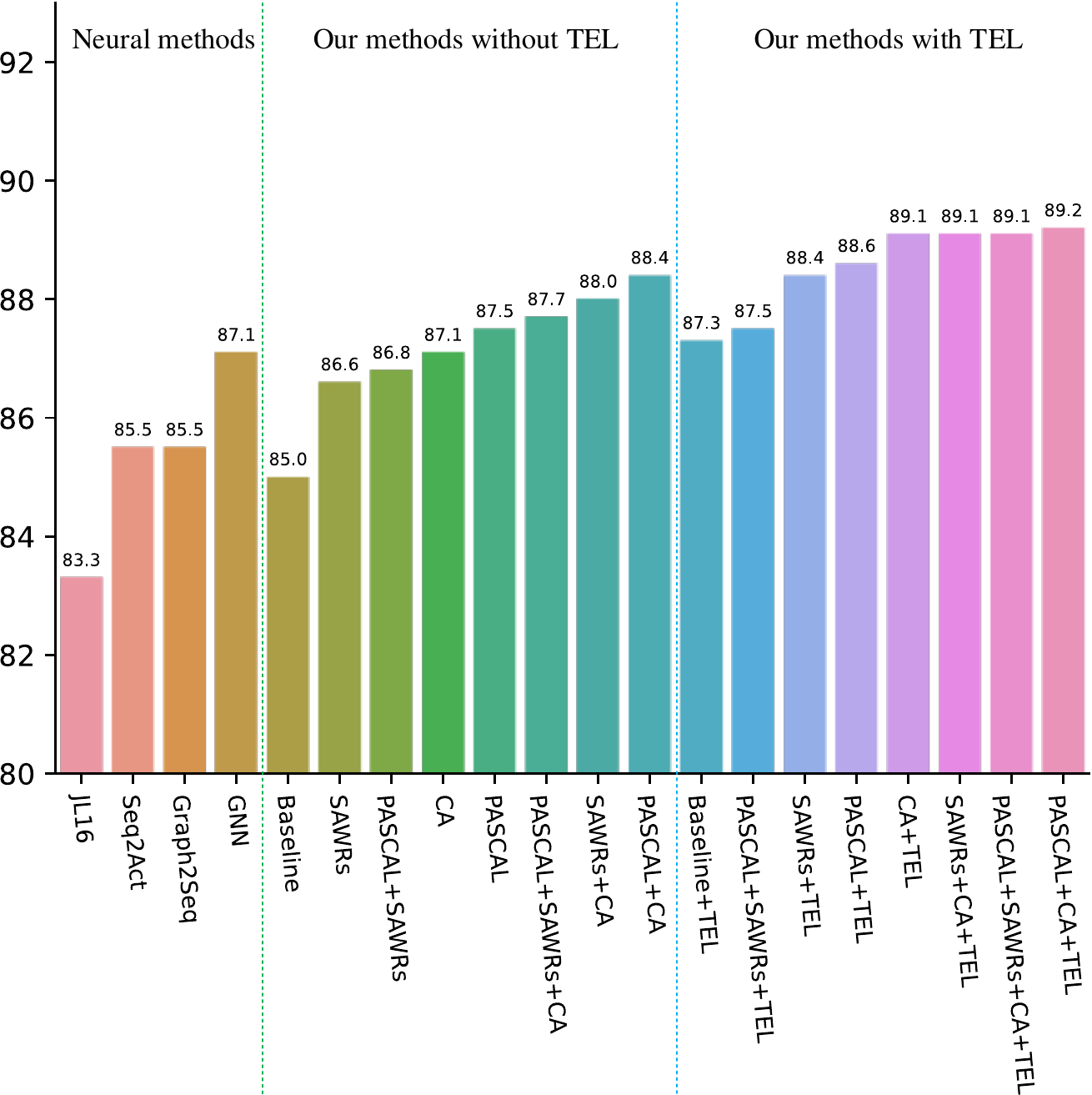}
			\label{ExactMatchonATIS}
		\end{minipage}
	}\hspace{-4mm}
	\quad
	\subfigure[{\scriptsize Tree Match on ATIS}]
	{
		\begin{minipage}{.18\linewidth}
			\centering
			\includegraphics[width=\linewidth]{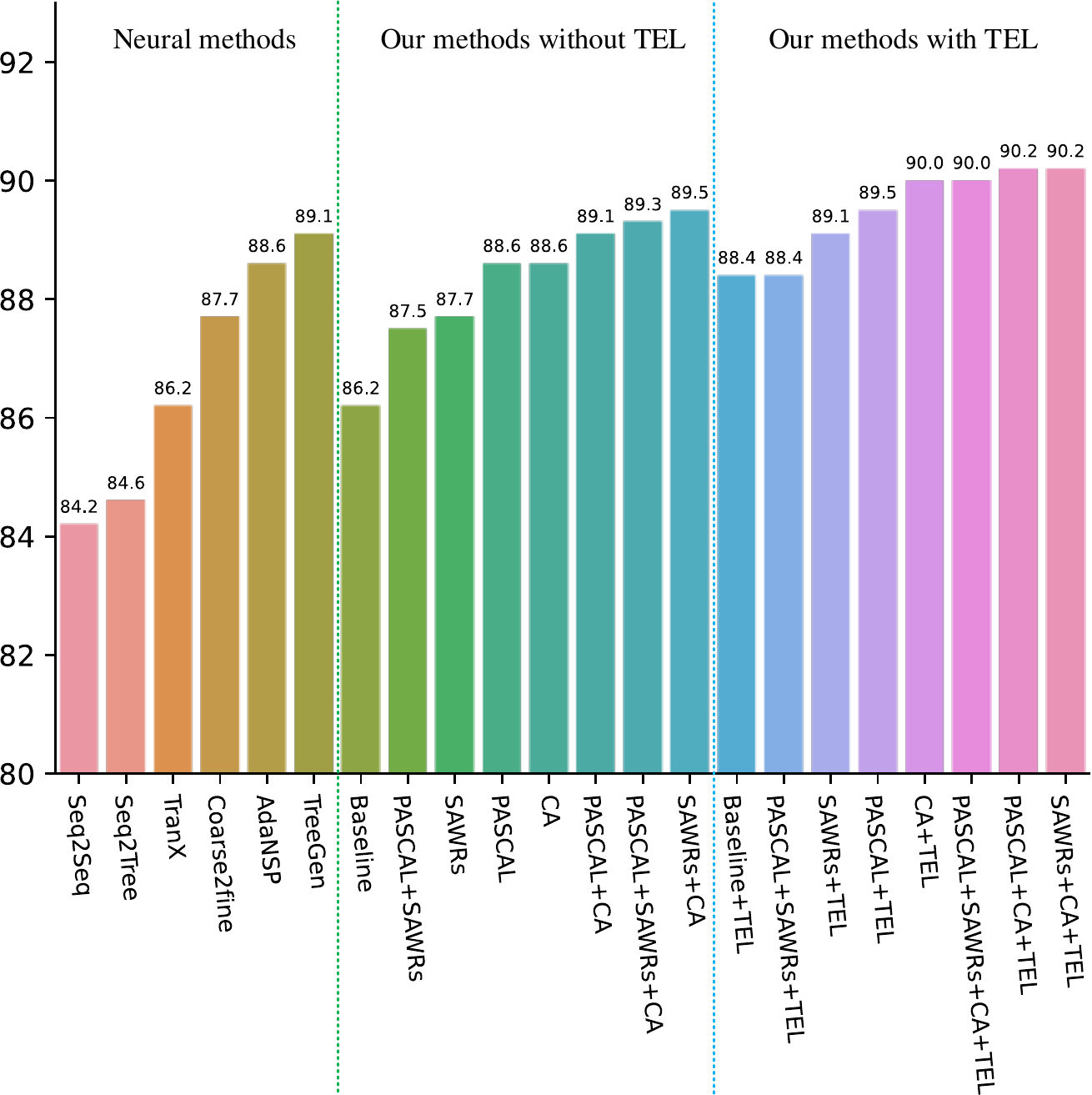}
			\label{TreeMatchonATIS}
		\end{minipage}
	}\hspace{-4mm}
	\quad
	\subfigure[{\scriptsize Exact Match on GEO}]
	{
		\begin{minipage}{.18\linewidth}
			\centering
			\includegraphics[width=\linewidth]{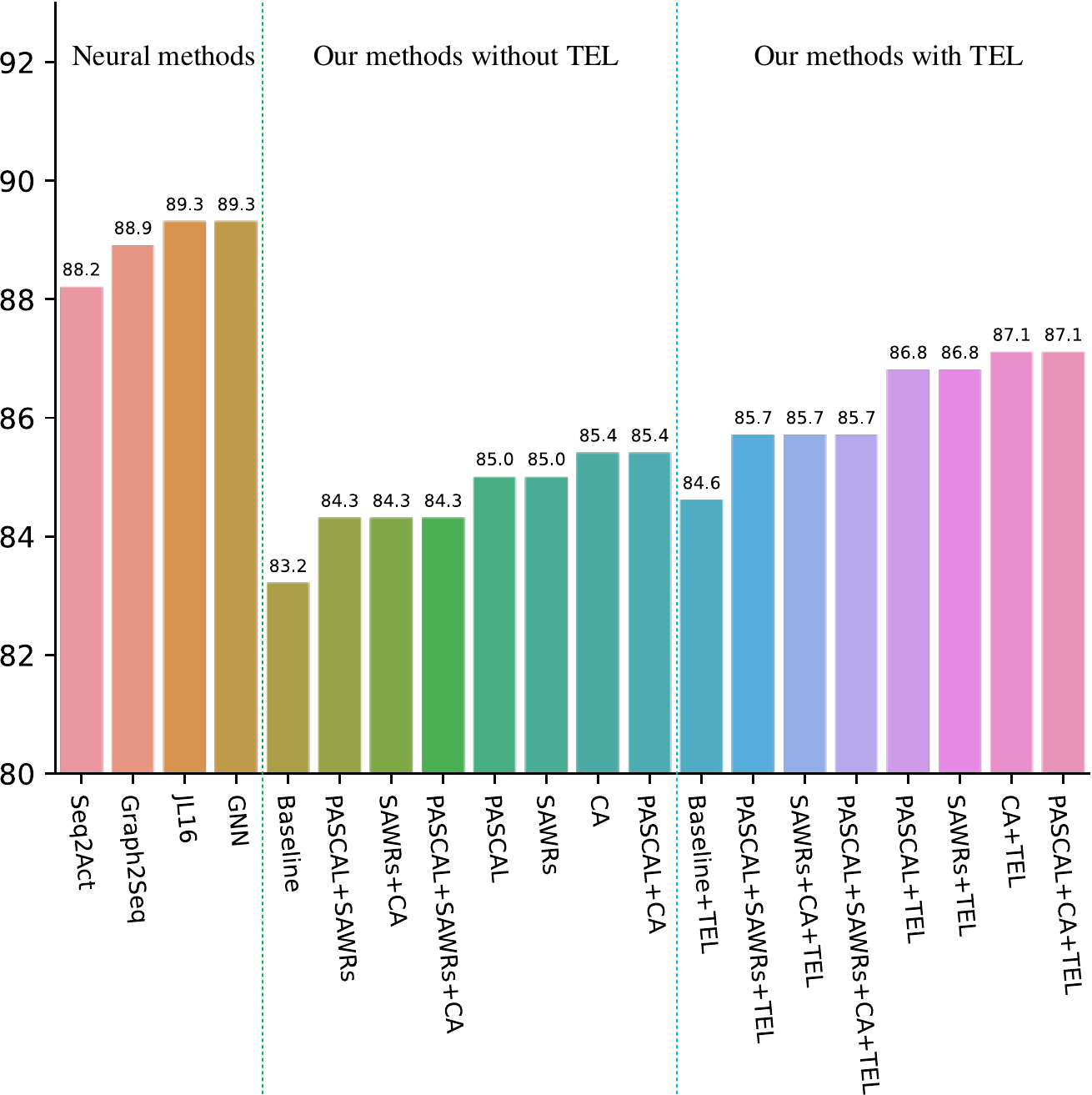}
			\label{ExactMatchonGEO}
		\end{minipage}
	}\hspace{-4mm}
	\quad
	\subfigure[{\scriptsize Tree Match on GEO}]
	{
		\begin{minipage}{.18\linewidth}
			\centering
			\includegraphics[width=\linewidth]{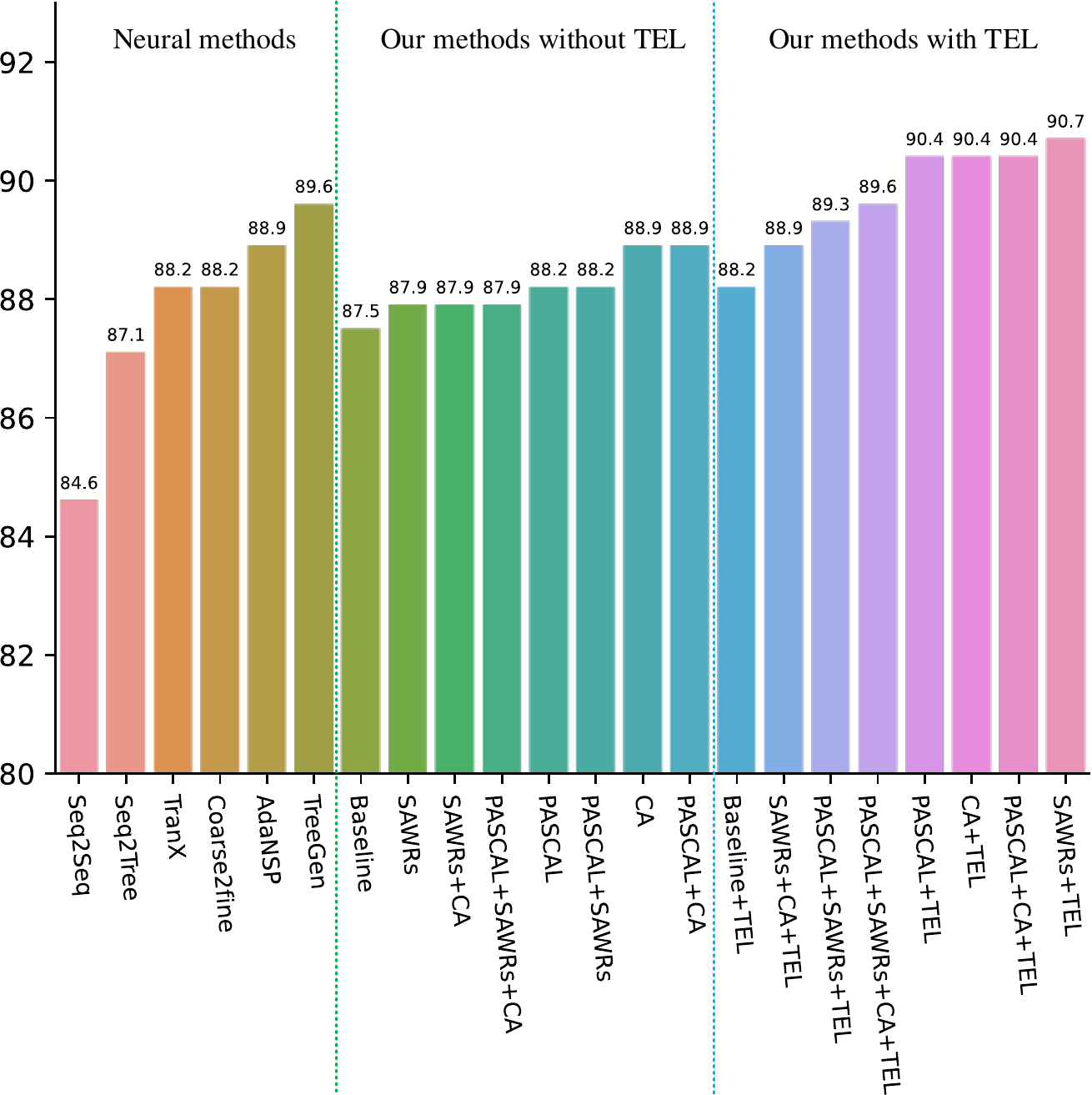}
			\label{TreeMatchonGEO}
		\end{minipage}
	}\hspace{-4mm}
	\quad
	\subfigure[{\scriptsize Exact Match on JOBS}]
	{
		\begin{minipage}{.18\linewidth}
			\centering
			\includegraphics[width=\linewidth]{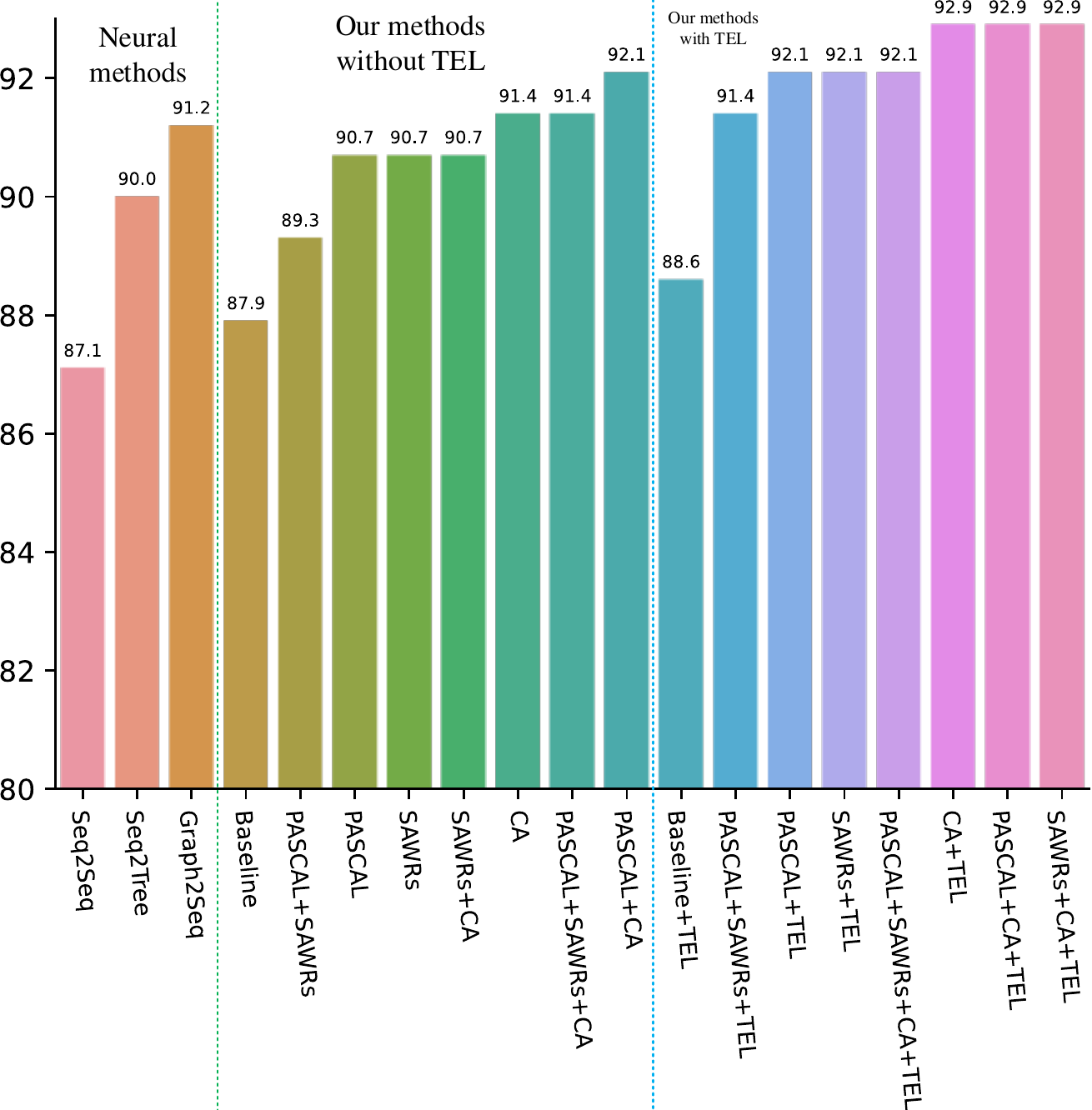}
			\label{ExactMatchonJOBS}
		\end{minipage}
	}
	\caption{Comparison of models on different datasets.}
	\label{result:barplot}
\end{figure}

\begin{table*}[htp]
	\centering
	\caption{\label{tab:size_time} Numbers of parameters and time costs per training epoch for our models.}
	\begin{tabular}{lcccc}
		\hline \multirow{2}{*}{\textbf{Model}} & \multirow{1}{*}{\textbf{Number of}} &  \multicolumn {3}{c} {\textbf{Time (s)}}   \\
		\cline{3-5} 
		& \textbf{Parameters (M)} & \textbf{ATIS} & \textbf{GEO} & \textbf{JOBS} \\
		\hline
		Baseline & 20.27 & 5.78 & 0.78 & 0.63 \\
		PASCAL & 20.27 & 7.77 & 0.99 & 0.83  \\
		SAWRs & 22.14 & 7.94 & 1.05 & 0.88  \\
		CA & 20.42 & 7.89 & 1.07 & 0.87  \\
		{PASCAL + SAWRs} & 22.14 & 9.69 & 1.19 & 1.03  \\
		{PASCAL + CA} & 20.42 & 9.06 & 1.10 & 0.96  \\
		{SAWRs + CA} & 22.29 & 9.49 & 1.26 & 1.06  \\
		{PASCAL + SAWRs + CA} & 22.29 & 11.75 & 1.45 & 1.25  \\
		\hline
	\end{tabular}
	
\end{table*}

\begin{figure*}[h]
	\centering
	\vspace{-0.35cm}
	\subfigtopskip=2pt
	\subfigbottomskip=0pt
	\subfigcapskip=-10pt
	\subfigure[Baseline]
	{
		\begin{minipage}{.18\linewidth}
			\centering
			\includegraphics[width=\linewidth]{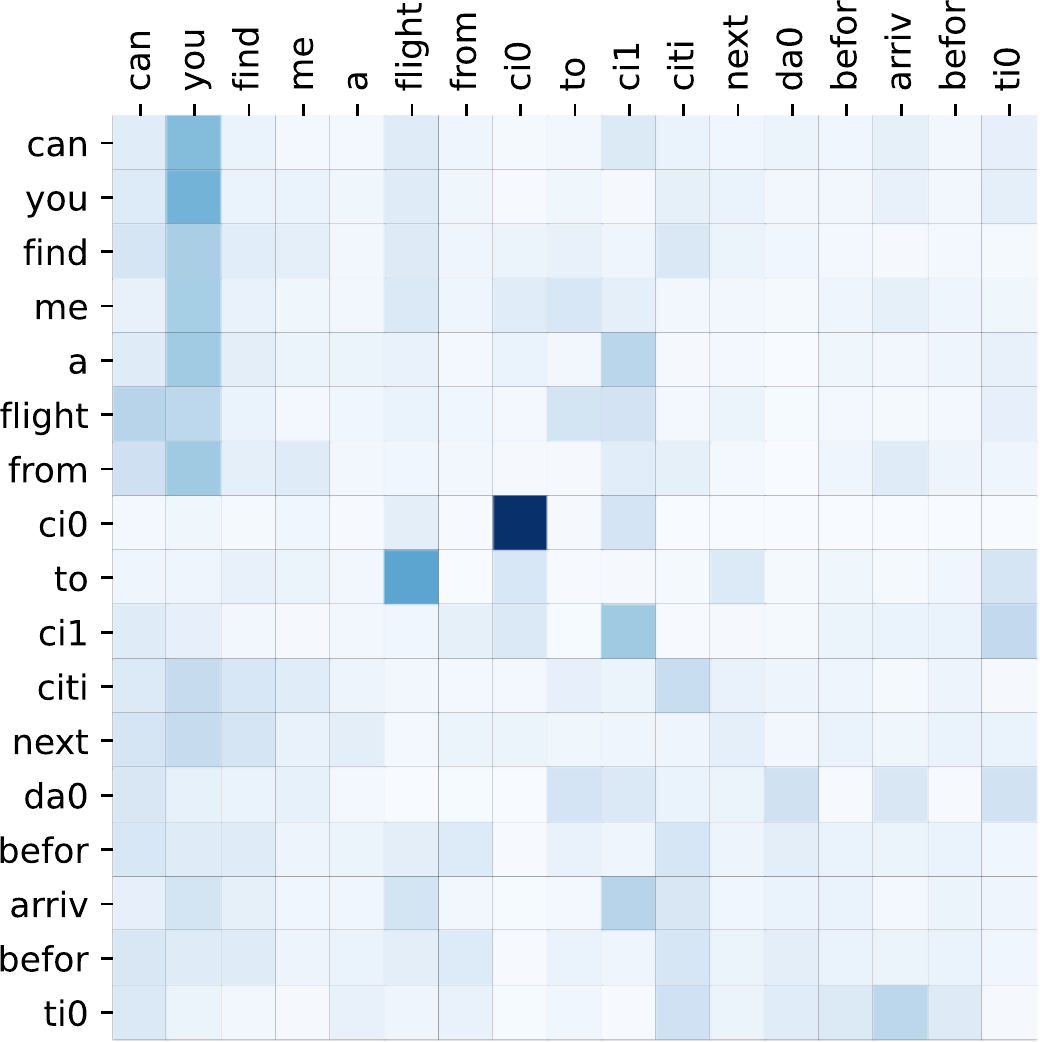}
			\label{atten1}
		\end{minipage}
	}\hspace{-2mm}
	\subfigure[SAWRs]
	{
		\begin{minipage}{.18\linewidth}
			\centering
			\includegraphics[width=\linewidth]{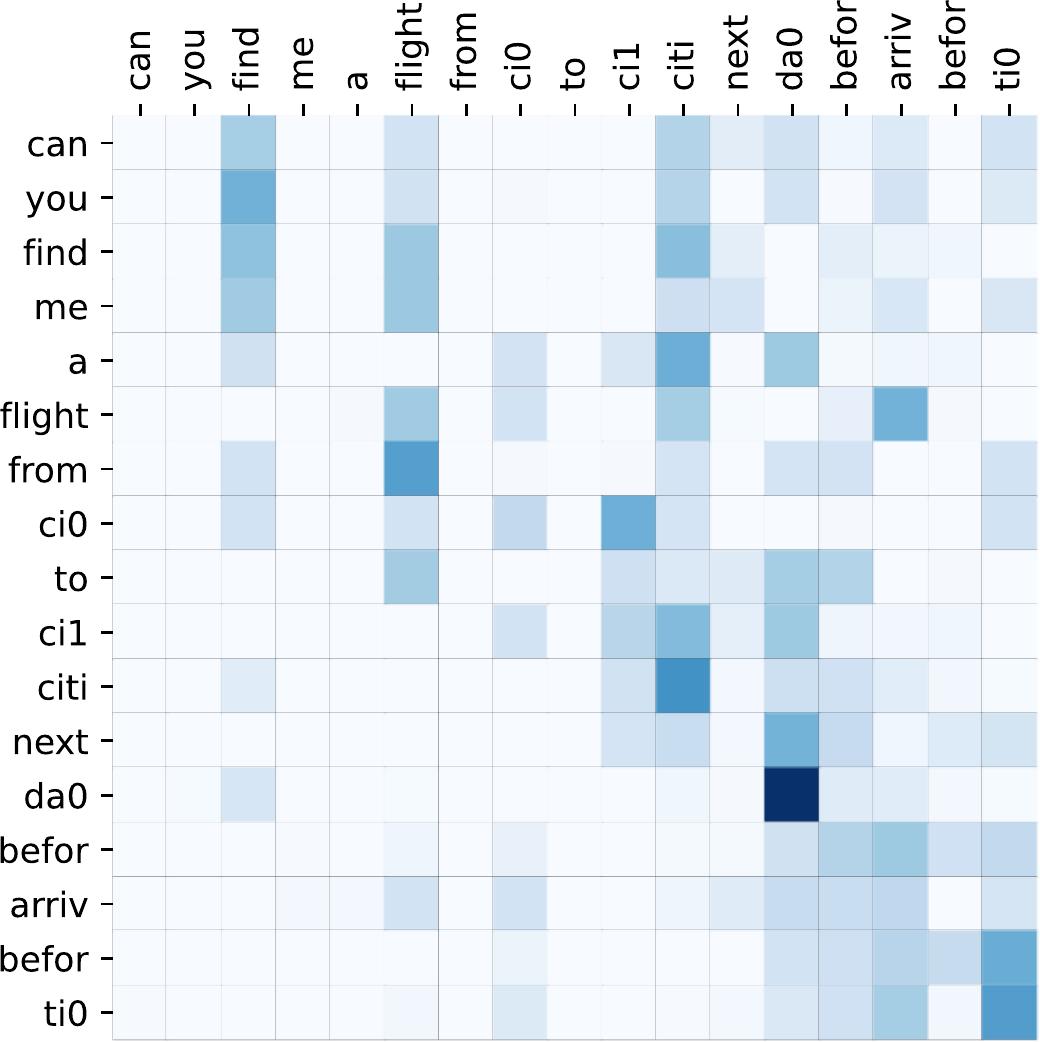}
			\label{atten2}
		\end{minipage}
	}\hspace{-2mm}
	\subfigure[CA]
	{
		\begin{minipage}{.18\linewidth}
			\centering
			\includegraphics[width=\linewidth]{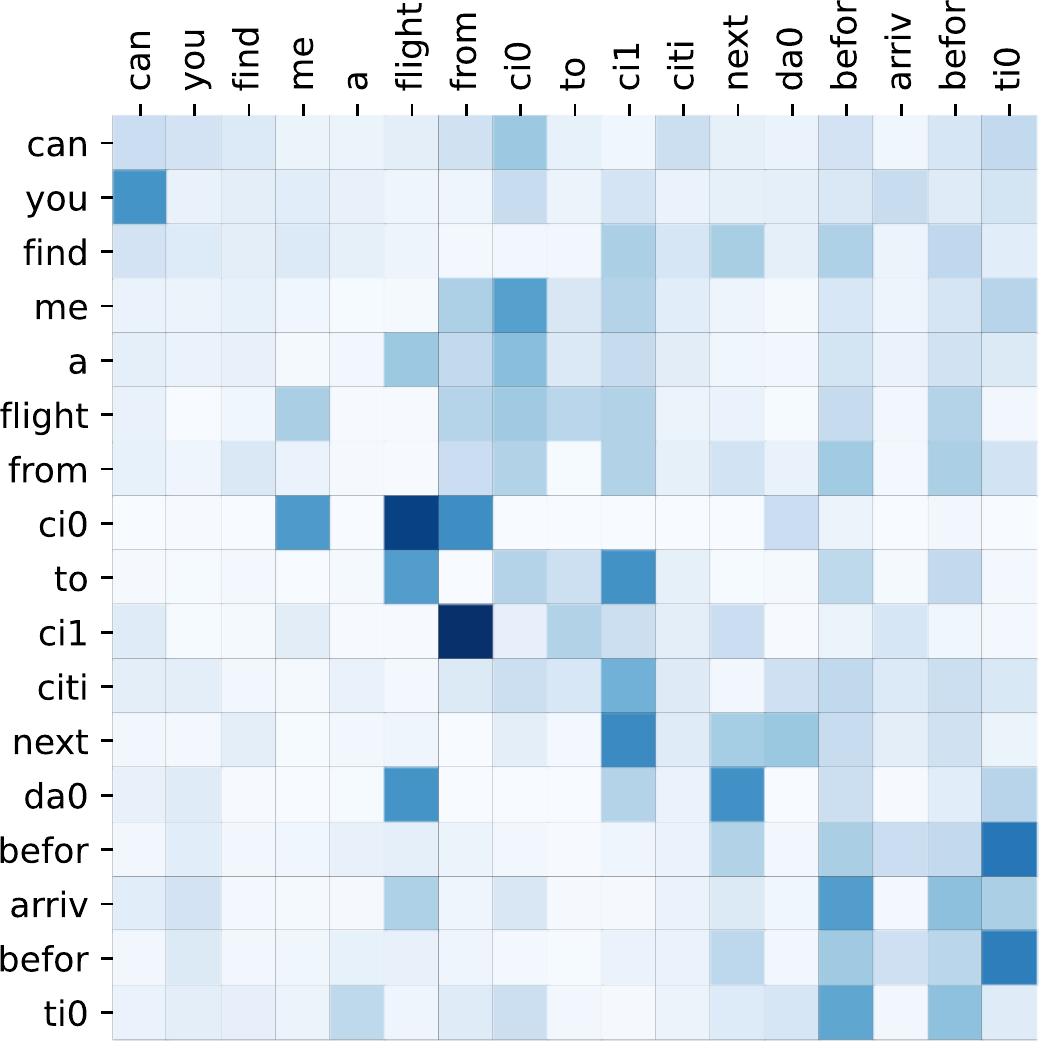}
			\label{atten3}
		\end{minipage}
	}\hspace{-2mm}
	\subfigure[PASCAL]
	{
		\begin{minipage}{.18\linewidth}
			\centering
			\includegraphics[width=\linewidth]{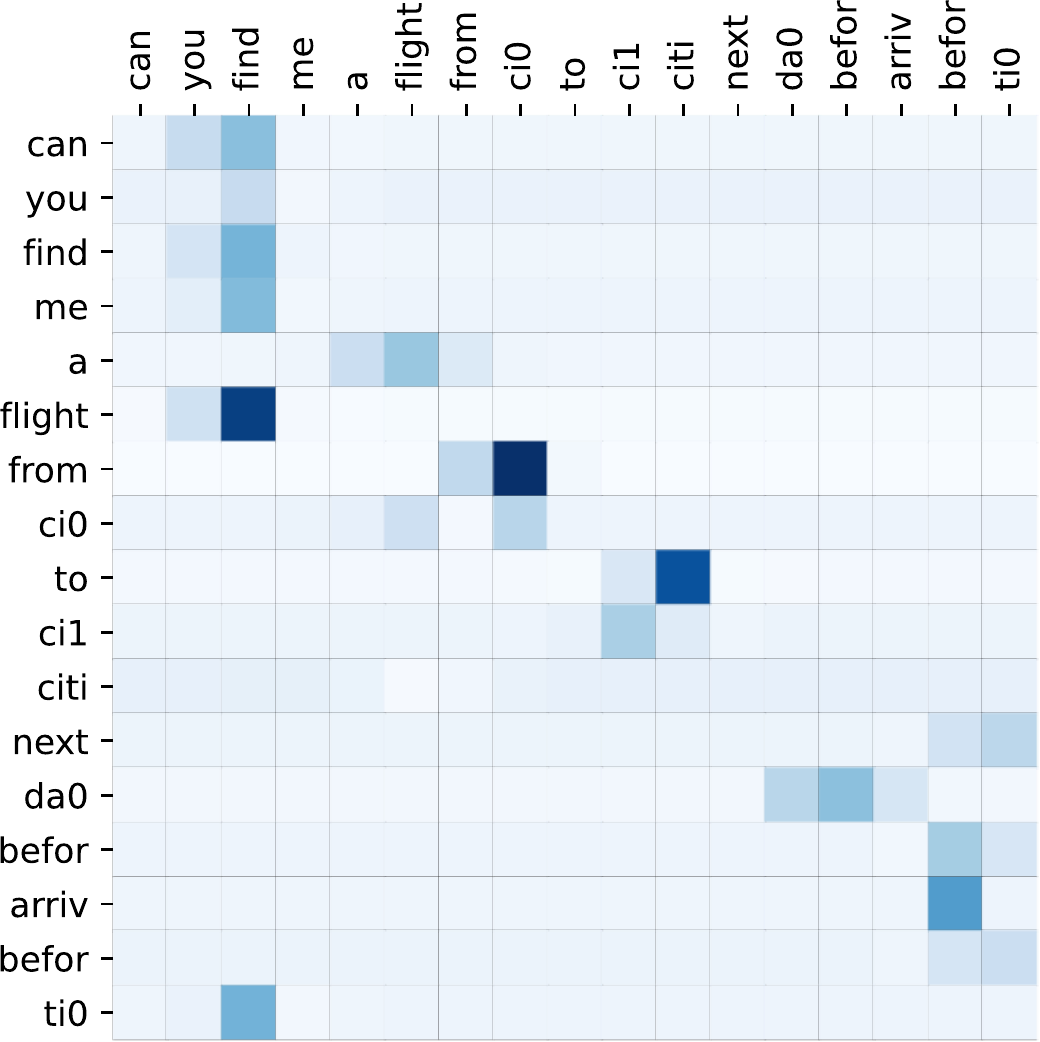}
			\label{atten4}
		\end{minipage}
	}\hspace{-2mm}
	\subfigure[$D$]
	{
		\begin{minipage}{.18\linewidth}
			\centering
			\includegraphics[width=\linewidth]{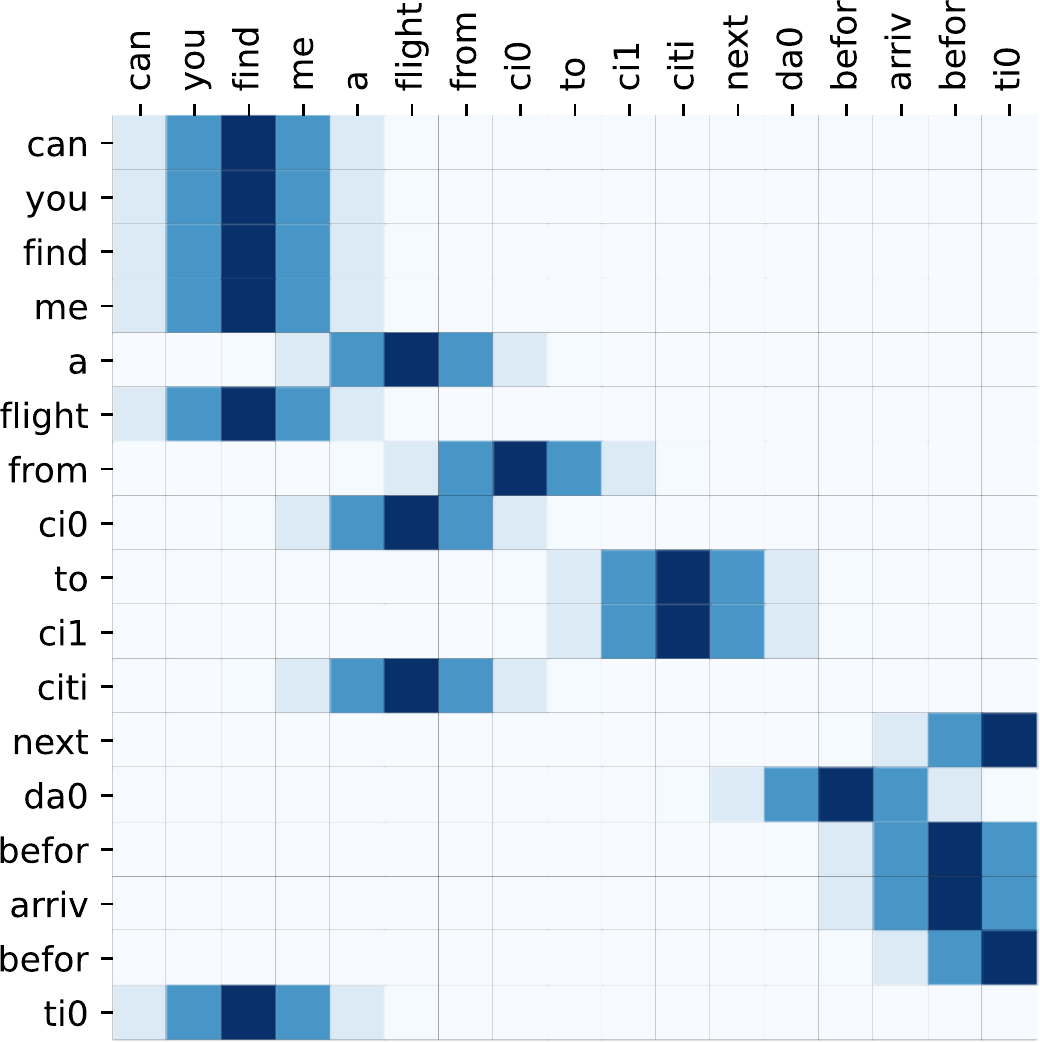}
			\label{atten4:parent}
		\end{minipage}
	}
	\caption{Heat maps of corresponding self-attention heads for each model and the 
		distance matrix $D$.}
	\label{atten}
\end{figure*}
\begin{figure*}[h]
	\centering
	\vspace{-0.35cm}
	\subfigtopskip=2pt
	\subfigbottomskip=0pt
	\subfigcapskip=-10pt
	\subfigure[CA]
	{
		\begin{minipage}{.23\linewidth}
			\centering
			\includegraphics[width=\linewidth]{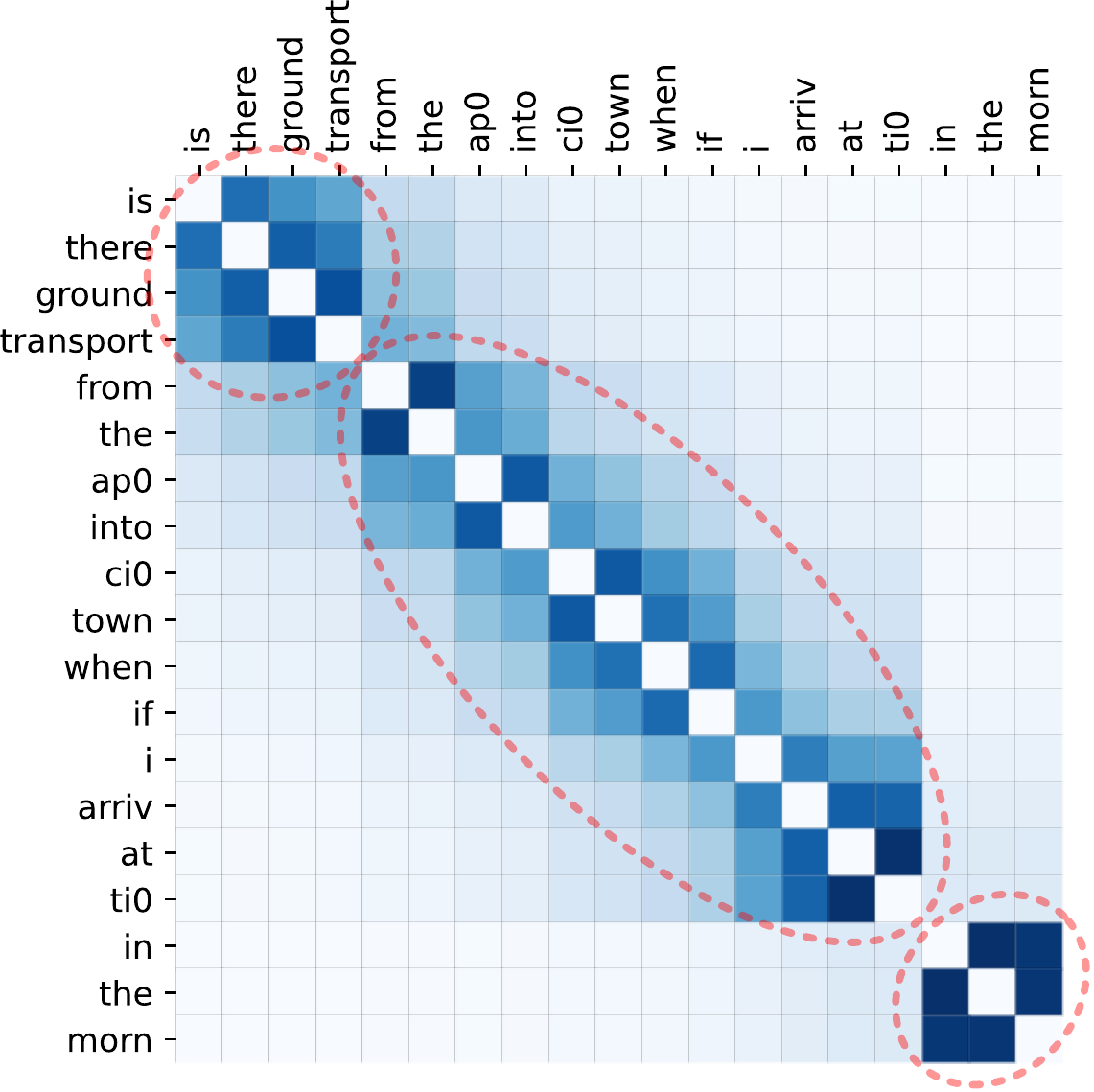}
			\label{treeca1}
		\end{minipage}
	}\hspace{-2mm}
	\subfigure[SAWRs + CA]
	{
		\begin{minipage}{.23\linewidth}
			\centering
			\includegraphics[width=\linewidth]{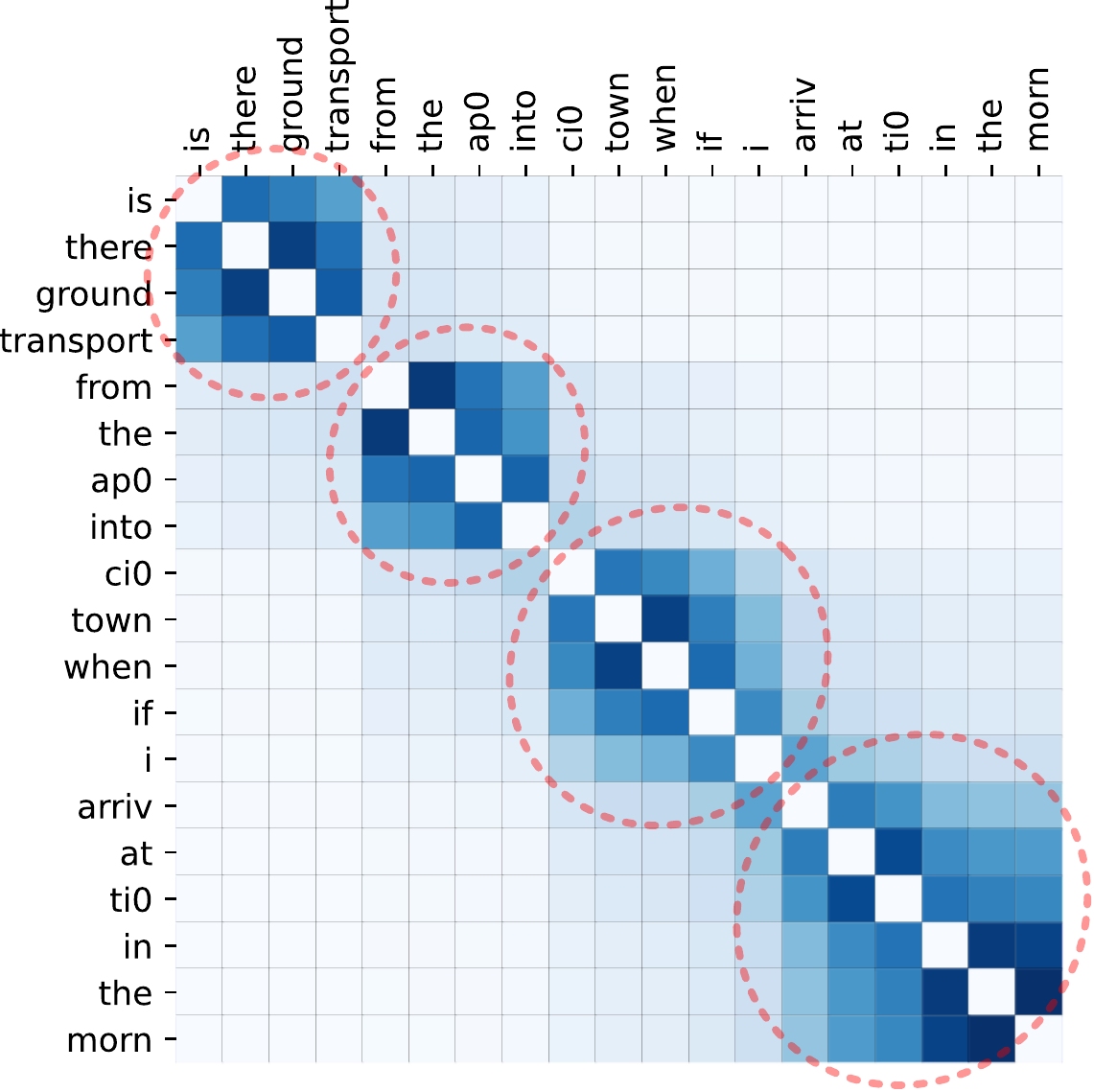}
			\label{treeca2}
		\end{minipage}
	}\hspace{-2mm}
	\subfigure[PASCAL + CA]
	{
		\begin{minipage}{.23\linewidth}
			\centering
			\includegraphics[width=\linewidth]{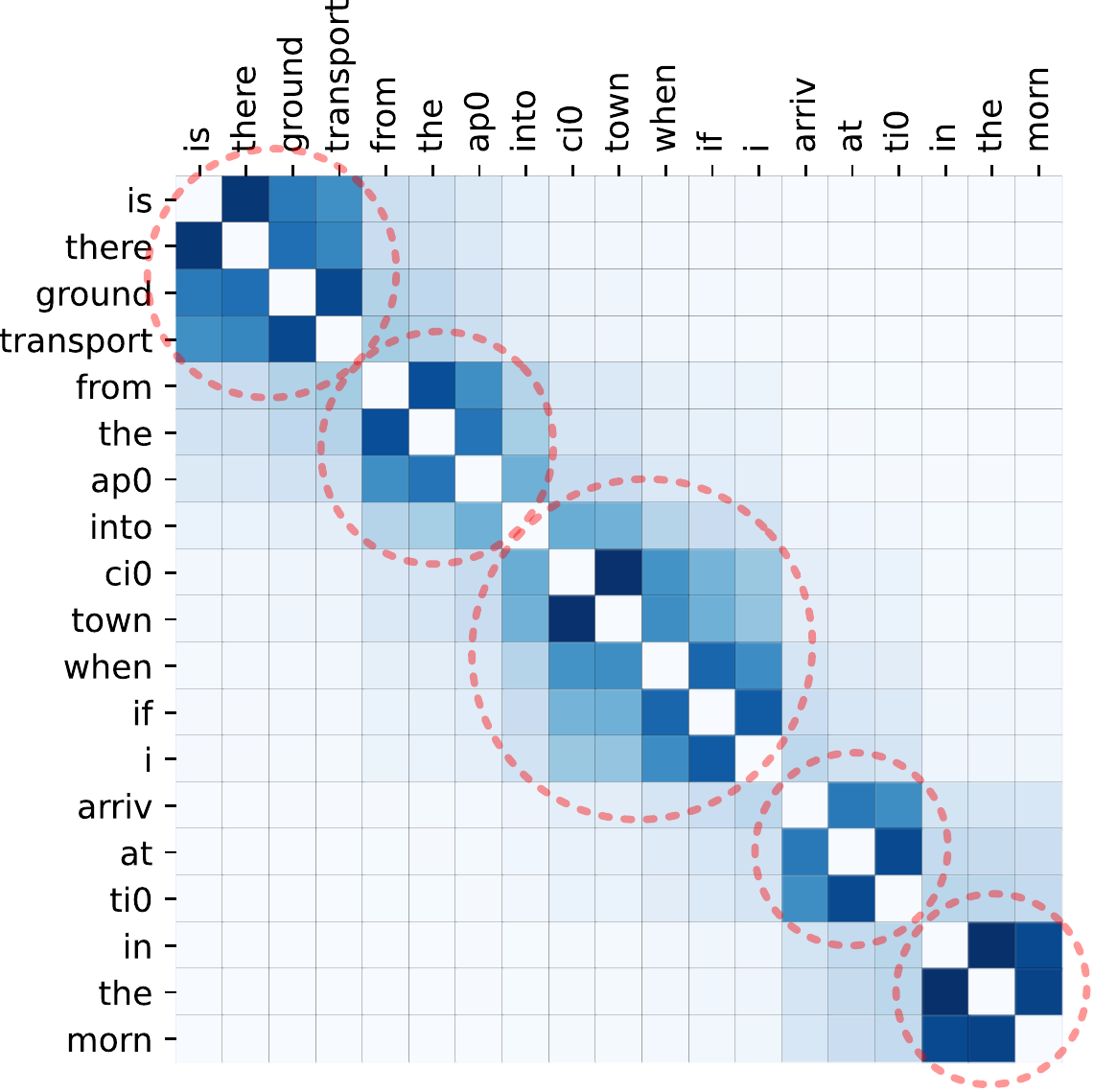}
			\label{treeca3}
		\end{minipage}
	}\hspace{-2mm}
	\subfigure[$D'$]
	{
		\begin{minipage}{.23\linewidth}
			\centering
			\includegraphics[width=\linewidth]{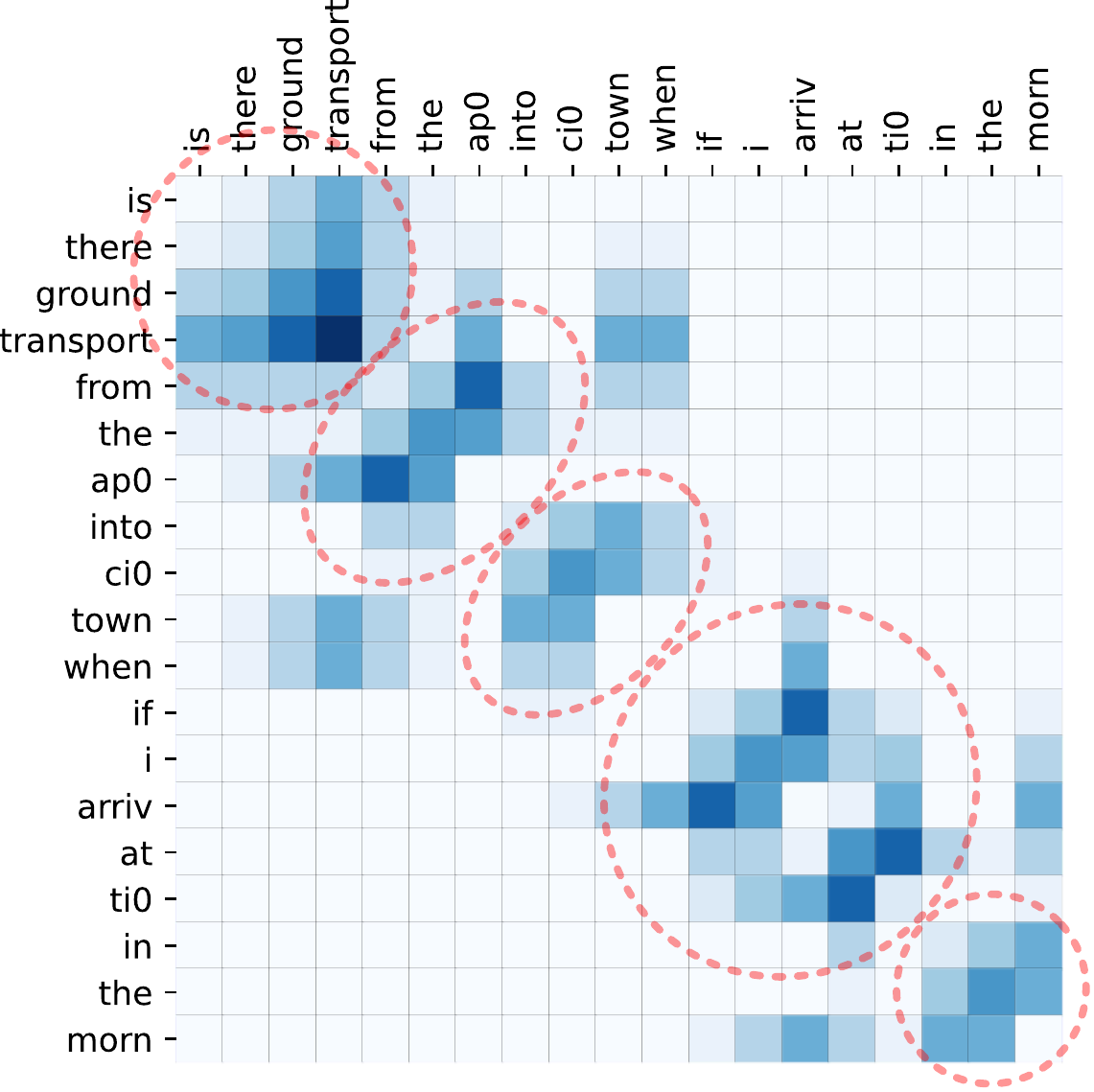}
			\label{treeca4}
		\end{minipage}
	}
	\caption{Heat maps of corresponding constituent prior $C$ for `CA', SAWRs + CA, PASCAL + CA, and $D'$.}
	\label{treeca}
\end{figure*}

Notice that, the three improvements and their combinations do not affect these hyperparameters for the baseline model. Then all the models considered in our experiments were trained based on such hyperparameters for their baseline part.

Networks in our experiments are implemented in PyTorch 
and trained with the AdamW optimizer 
with its default parameters. 
We trained every model for 45 (resp. 250) epochs for ATIS (resp. GEO and JOBS) on two GPUs, i.e., Nvidia GeForce 2080Ti and Nvidia RTX 3090, which takes around 2 (resp. 3) hours for ATIS (resp. GEO and JOBS).

In SAWRs, a neural dependency parser needs to be pre-trained to obtain its intermediate hidden representations. In this paper, we implement such a dependency parser based on the model in~\cite{simperparser}. 
We first adopt Stanford CoreNLP package~\cite{stanfordparser} to obtain the dependency trees and the corresponding part-of-speech tagging sequences.
Then we use these dependency trees as labels for these sequences and train the dependency parser based on this synthetic dataset. it eliminates the need of manual work for labeling dependency data of ATIS,GEO or JOBS.

We use AdamW as the optimizer and set the learning rate as 1e-4 for the training. We applied the ensemble learning method TEL to further improve the performance of models. 
We also tried the conventional ensemble method~\cite{hansen1990neural}. However, it did not perform well in our experiments.

\subsection{Results}
\label{sec:result}

Table~\ref{tab:result} summarizes the performance of our models in datasets, where `Baseline' denotes the baseline model without any improvements, and `PASCAL' (reps. `SAWRs' and `CA') denotes the model that applies the improvement PASCAL (resp. SAWRs and CA) on the baseline model.
We also illustrate the results in Figure~\ref{result:barplot}. 

The experimental results show that all three improvements can improve the performance of the baseline model. 
Although the baseline model is very simple, the models `PASCAL', `SAWRs', and `CA' still achieve good performance among other neural methods. 
Moreover, the performance can be further improved by applying TEL. 
These results show that the improvements PASCAL, SAWRs, and CA are effective in exploiting dependency information for semantic parsing, and TEL can further improve the performance of a semantic parser. 

For the combinations of the three improvements, we find out that PASCAL + CA and SAWRs + CA achieve better performance than PASCAL + SAWRs. 
PASCAL + SAWRs performs better than `Baseline'. However, it even performs worse than the model `PASCAL' or `SAWRs' in some cases. 
It seems that both PASCAL and SAWRs obtain similar information from the dependency tree. Then the combination of both does not provide additional benefits. 
On the other hand, PASCAL + CA, SAWRs + CA, PASCAL + SAWRs + CA achieve good performance among all neural methods. 
This implies that the dependency information obtained from CA is complementary to the one from PASCAL or SAWRs. 
PASCAL + CA is much simpler than SAWRs + CA and PASCAL + SAWRs + CA. Then we suggests applying the combination of PASCAL and CA on the Transformer based seq2seq semantic parser.
The performance of these combinations can also be further improved by applying TEL. 
These results show that CA is complementary to PASCAL and SAWRs, PASCAL + CA provides state-of-the-art performance among neural methods, and TEL can further improve the performance. 

Notice that, our methods do not perform well for Exact Match on GEO. This is mainly due to the facts that the size of GEO is small and Exact Match causes spurious errors.
We can observe that our methods perform well for Tree Match on GEO.

We also compare the training costs of the three improvements and their combinations.
Table~\ref{tab:size_time} summarizes numbers of parameters and time costs per training epoch on ATIS for our models.
Note that, the size of `PASCAL', `CA', or PASCAL + CA is almost the same as `Baseline'. `SAWRs' and the combinations with it introduce a few additional parameters for `Baseline'.
The time cost per training epoch is slightly increased when an improvement is applied to `Baseline' and a combination requires more time.
Notice that, PASCAL + CA achieves state-of-the-art performance with the similar size of `Baseline' and a slight increase in time cost.

\subsection{Visual Analysis}

In this section, we try to analyze what has been learned to help the baseline model by applying each improvement and what has been learned for CA and combinations with it.

We first visualize the heat map of the weights on $V$, i.e., the attention weights of tokens for the input sentence, which are obtained from the self-attention heads of the last layer in the encoder for each model. 
Figure~\ref{atten} illustrates such heat maps for `Baseline', `SAWRs', `CA', and `PASCAL'. Figure~\ref{atten4:parent} also illustrates the heat map of the distance matrix $D$ obtained from the dependency tree of the sentence, as specified in Section~\ref{sec:PASCAL}.
Notice that, the heat map for `PASCAL' is similar to the one of $D$ due to Equation~\eqref{eq:pascal}. 
We can also observe that, different from the heat map for `Baseline', the ones for `SAWRs' and `CA' are more similar to the heat map for `PASCAL'. 
This implies that both SAWRS and CA tend to encourage the self-attention heads to follow the structure of the dependency tree.

On the other hand, we also visualize the heat map of the constituent prior $C$ used in CA. 
Figure~\ref{treeca} illustrates heat maps of corresponding constituent priors for `CA', SWARs + CA, and PASCAL + CA. 
Notice that, $C$ is a symmetrical matrix. 
Then we can generate a symmetrical matrix $D'$ from the distance matrix $D$ of the sentence by $D' = \frac{D+D^\top}{2}$.
Figure~\ref{treeca4} also illustrates the heat map of $D'$. 
We can observe that, the heat maps for PASCAL + CA and SWARs + CA are more and more similar to the one of $D'$. 
This implies that the combinations PASCAL + CA and SWARs + CA help the constituent prior $C$ to capture information from the dependency tree of the sentence, and PASCAL + CA captures more information.
This observation partially explains the reason why PASCAL + CA performs better.

\section{Conclusion}

In this paper, we implement three improvements, i.e., PASCAL, SAWRs, and CA, to incorporate the dependency information of input sentences in a Transformer encoder for a seq2seq semantic parser. 
We show that all three improvements are effective in exploiting such dependency information for semantic parsing with a slight increase in training cost. 
We also examine the combinations of these improvements. 
We observe that both PASCAL and SAWRs obtain similar information from the dependency tree, and the combination of both does not provide additional benefits. 
We find out that CA is complementary to PASCAL and SAWRs, and PASCAL + CA provides state-of-the-art performance among neural approaches on ATIS, GEO, and JOBS datasets.
Moreover, PASCAL + CA can be implemented easily with a slight increase in training costs.
We provide visual analysis that tries to explain why PASCAL + CA performs better among other improvements and combinations.
We also implement TEL for the models and show that TEL is effective for these improvements on semantic parsing.

\section{Acknowledgment}
The work is partially supported by the National Key Research and Development Program of China (No. 2018AAA0100500), CAAI-Huawei MindSpore Open Fund, Anhui Provincial Development and Reform Commission 2020 New Energy Vehicle Industry Innovation Development Project ``Key System Research and Vehicle Development for Mass Production Oriented Highly Autonomous Driving'', and Key-Area Research and Development Program of Guangdong Province 2020B0909050001.

\bibliographystyle{splncs04}
\bibliography{myreference}

\begin{thebibliography}{10}
\providecommand{\url}[1]{\texttt{#1}}
\providecommand{\urlprefix}{URL }
\providecommand{\doi}[1]{https://doi.org/#1}

\bibitem{bahdanau2015neural}
Bahdanau, D., Cho, K.H., Bengio, Y.: Neural machine translation by jointly
  learning to align and translate. In: Proceedings of the 3rd International
  Conference on Learning Representations (2015)

\bibitem{enhancingMTwithDASA}
Bugliarello, E., Okazaki, N.: Enhancing machine translation with
  dependency-aware self-attention. In: Proceedings of the 58th Annual Meeting
  of the Association for Computational Linguistics. pp. 1618--1627. Association
  for Computational Linguistics (2020)

\bibitem{chen2018sequence}
Chen, B., Sun, L., Han, X.: Sequence-to-action: End-to-end semantic graph
  generation for semantic parsing. In: Proceedings of the 56th Annual Meeting
  of the Association for Computational Linguistics. pp. 766--777. Association
  for Computational Linguistics (2018)

\bibitem{cho2014learning}
Cho, K., van Merrienboer, B., Gulcehre, C., Bougares, F., Schwenk, H., Bengio,
  Y.: Learning phrase representations using rnn encoder-decoder for statistical
  machine translation. In: Proceedings of the 2014 Conference on Empirical
  Methods in Natural Language Processing (2014)

\bibitem{dong2016language}
Dong, L., Lapata, M.: Language to logical form with neural attention. In:
  Proceedings of the 54th Annual Meeting of the Association for Computational
  Linguistics. pp. 33--43. Association for Computational Linguistics (2016)

\bibitem{dong2018coarse}
Dong, L., Lapata, M.: Coarse-to-fine decoding for neural semantic parsing. In:
  Proceedings of the 56th Annual Meeting of the Association for Computational
  Linguistics. pp. 731--742. Association for Computational Linguistics (2018)

\bibitem{deepbiaffparser}
Dozat, T., Manning, C.D.: Deep biaffine attention for neural dependency
  parsing. In: Proceedings of the 5th International Conference on Learning
  Representations (2017)

\bibitem{simperparser}
Dozat, T., Manning, C.D.: Simpler but more accurate semantic dependency
  parsing. In: Proceedings of the 56th Annual Meeting of the Association for
  Computational Linguistics. pp. 484--490. Association for Computational
  Linguistics (2018)

\bibitem{hansen1990neural}
Hansen, L.K., Salamon, P.: Neural network ensembles. IEEE transactions on
  pattern analysis and machine intelligence  \textbf{12}(10),  993--1001 (1990)

\bibitem{jia-liang-2016-data}
Jia, R., Liang, P.: Data recombination for neural semantic parsing. In:
  Proceedings of the 54th Annual Meeting of the Association for Computational
  Linguistics. pp. 12--22. Association for Computational Linguistics (2016)

\bibitem{kwiatkowski-etal-2013-scaling}
Kwiatkowski, T., Choi, E., Artzi, Y., Zettlemoyer, L.: Scaling semantic parsers
  with on-the-fly ontology matching. In: Proceedings of the 2013 Conference on
  Empirical Methods in Natural Language Processing. pp. 1545--1556 (2013)

\bibitem{kwiatkowski2011lexical}
Kwiatkowski, T., Zettlemoyer, L., Goldwater, S., Steedman, M.: Lexical
  generalization in {CCG} grammar induction for semantic parsing. In:
  Proceedings of the 2011 Conference on Empirical Methods in Natural Language
  Processing. pp. 1512--1523. Association for Computational Linguistics (2011)

\bibitem{liang39learning}
Liang, P., Jordan, M.I., Klein, D.: Learning dependency-based compositional
  semantics. Computational Linguistics  \textbf{39}(2),  389--446 (2013)

\bibitem{stanfordparser}
Manning, C.D., Surdeanu, M., Bauer, J., Finkel, J.R., Bethard, S., McClosky,
  D.: The stanford {CoreNLP} natural language processing toolkit. In:
  Proceedings of 52nd Annual Meeting of the Association for Computational
  Linguistics: System Demonstrations. pp. 55--60. Association for Computational
  Linguistics (2014)

\bibitem{dependencytree1}
Reddy, S., T{\"a}ckstr{\"o}m, O., Collins, M., Kwiatkowski, T., Das, D.,
  Steedman, M., Lapata, M.: Transforming dependency structures to logical forms
  for semantic parsing. Transactions of the Association for Computational
  Linguistics  \textbf{4},  127--140 (2016)

\bibitem{dependencytree2}
Reddy, S., T{\"a}ckstr{\"o}m, O., Petrov, S., Steedman, M., Lapata, M.:
  Universal semantic parsing. In: Proceedings of the 2017 Conference on
  Empirical Methods in Natural Language Processing. pp. 89--101 (2017)

\bibitem{shaw2019generating}
Shaw, P., Massey, P., Chen, A., Piccinno, F., Altun, Y.: Generating logical
  forms from graph representations of text and entities. In: Proceedings of the
  57th Annual Meeting of the Association for Computational Linguistics. pp.
  95--106. Association for Computational Linguistics (2019)

\bibitem{sun2019grammar}
Sun, Z., Zhu, Q., Mou, L., Xiong, Y., Li, G., Zhang, L.: A grammar-based
  structural cnn decoder for code generation. In: Proceedings of the 33rd AAAI
  Conference on Artificial Intelligence. pp. 7055--7062 (2019)

\bibitem{sun2020treegen}
Sun, Z., Zhu, Q., Xiong, Y., Sun, Y., Mou, L., Zhang, L.: {TreeGen}: A
  tree-based transformer architecture for code generation. In: Proceedings of
  the 34th AAAI Conference on Artificial Intelligence. pp. 8984--8991 (2020)

\bibitem{vaswani2017attention}
Vaswani, A., Shazeer, N., Parmar, N., Uszkoreit, J., Jones, L., Gomez, A.N.,
  Kaiser, {\L}., Polosukhin, I.: Attention is all you need. In: Advances in
  Neural Information Processing Systems. pp. 5998--6008. MIT Press (2017)

\bibitem{wang2014morpho}
Wang, A., Kwiatkowski, T., Zettlemoyer, L.: Morpho-syntactic lexical
  generalization for {CCG} semantic parsing. In: Proceedings of the 2014
  Conference on Empirical Methods in Natural Language Processing. pp.
  1284--1295. Association for Computational Linguistics (2014)

\bibitem{Wang2019TreeTI}
Wang, Y.S., Lee, H.Y., Chen, Y.N.: Tree transformer: Integrating tree
  structures into self-attention. In: Proceedings of the 2019 Conference on
  Empirical Methods in Natural Language Processing and the 9th International
  Joint Conference on Natural Language Processing. pp. 1061--1070 (2019)

\bibitem{transductiveEL}
Wang, Y., Wu, L., Xia, Y., Qin, T., Zhai, C., Liu, T.Y.: Transductive ensemble
  learning for neural machine translation. In: Proceedings of The 34th AAAI
  Conference on Artificial Intelligence. pp. 6291--6298 (2020)

\bibitem{xu2018exploiting}
Xu, K., Wu, L., Wang, Z., Yu, M., Chen, L., Sheinin, V.: Exploiting rich
  syntactic information for semantic parsing with graph-to-sequence model. In:
  Proceedings of the 2018 Conference on Empirical Methods in Natural Language
  Processing. pp. 918--924. Association for Computational Linguistics (2018)

\bibitem{Yin2018TRANXAT}
Yin, P., Neubig, G.: {TRANX}: A transition-based neural abstract syntax parser
  for semantic parsing and code generation. In: Proceedings of the 2018
  Conference on Empirical Methods in Natural Language Processing. pp. 7--12.
  Association for Computational Linguistics (2018)

\bibitem{zettlemoyer2007online}
Zettlemoyer, L.S., Collins, M.: Online learning of relaxed {CCG} grammars for
  parsing to logical form. In: Proceedings of the 2007 Joint Conference on
  Empirical Methods in Natural Language Processing and Computational Natural
  Language Learning. pp. 678--687 (2007)

\bibitem{syntaxenhancedNMT}
Zhang, M., Li, Z., Fu, G., Zhang, M.: Syntax-enhanced neural machine
  translation with syntax-aware word representations. In: Proceedings of the
  2019 Conference of the North American Chapter of the Association for
  Computational Linguistics: Human Language Technologies. pp. 1151--1161 (2019)

\bibitem{zhang2019adansp}
Zhang, X., He, S., Liu, K., Zhao, J.: {AdaNSP}: Uncertainty-driven adaptive
  decoding in neural semantic parsing. In: Proceedings of the 57th Annual
  Meeting of the Association for Computational Linguistics. pp. 4265--4270.
  Association for Computational Linguistics (2019)

\bibitem{zhao2015type}
Zhao, K., Huang, L.: Type-driven incremental semantic parsing with
  polymorphism. In: Proceedings of the 2015 Conference of the North American
  Chapter of the Association for Computational Linguistics: Human Language
  Technologies. pp. 1416--1421 (2015)

\end{thebibliography}
%
%
%
%
\end{document}